\documentclass[iicol,sn-mathphys-num]{sn-jnl}% Math and Physical Sciences Numbered Reference Style

\usepackage[utf8]{inputenc}   % UTF-8 input
\usepackage[T1]{fontenc}      % 8-bit T1 fonts

\usepackage{amsmath}
\usepackage{amssymb}
\usepackage{amsfonts}
\usepackage{bm}               % Bold math symbols
\usepackage{mathrsfs}
\usepackage{dsfont}
\usepackage{nicefrac}         % Compact symbols for 1/2 etc.

\usepackage{graphicx}
\usepackage{tikz}
\usepackage{xcolor}
\usepackage{colortbl}
\definecolor{lightpurple}{RGB}{224, 224, 250}
\definecolor{dg}{rgb}{0,0.694,0.298}
\definecolor{purple}{rgb}{0.4,0.176,0.569}
\definecolor{royalblue}{RGB}{65,105,225}
\usepackage{graphicx}

\usepackage{booktabs}
\usepackage{multirow}
\usepackage{tabularx}
\usepackage{longtable}
\usepackage{makecell}
\usepackage{caption}
\usepackage[font=footnotesize,labelfont=bf]{caption}
\usepackage[belowskip=-5pt,aboveskip=3pt]{caption}
\usepackage{enumitem}
\usepackage{pdflscape}
\usepackage{algorithm}
\usepackage{algorithmicx}
\usepackage{algpseudocode}
\usepackage{algcompatible}

\algnewcommand\INPUT{\item[\textbf{Input:}]}
\algnewcommand\OUTPUT{\item[\textbf{Output:}]}
\newbox\statebox
\newcommand{\myState}[1]{%
    \setbox\statebox=\vbox{#1}%
    \edef\thealgruleheight{\dimexpr \the\ht\statebox+1pt\relax}%
    \edef\thealgruledepth{\dimexpr \the\dp\statebox+1pt\relax}%
    \ifdim\thealgruleheight<.75\baselineskip
        \def\thealgruleheight{\dimexpr .75\baselineskip+1pt\relax}%
    \fi
    \ifdim\thealgruledepth<.25\baselineskip
        \def\thealgruledepth{\dimexpr .25\baselineskip+1pt\relax}%
    \fi
    \State #1%
    \def\thealgruleheight{\dimexpr .75\baselineskip+1pt\relax}%
    \def\thealgruledepth{\dimexpr .25\baselineskip+1pt\relax}%
}

\usepackage{listings}
\usepackage{pifont}

\usepackage{url}
\usepackage{comment}
\usepackage{sidecap}
\usepackage{wrapfig}
\usepackage[title]{appendix}
\usepackage{textcomp}
\usepackage[accsupp]{axessibility}  % Improves PDF accessibility

\usepackage{amsthm}
\theoremstyle{thmstyleone}%
\theoremstyle{thmstyletwo}%
\theoremstyle{thmstylethree}%
\usepackage{xspace}
\makeatletter
\DeclareRobustCommand\onedot{\futurelet\@let@token\@onedot}
\def\@onedot{\ifx\@let@token.\else.\null\fi\xspace}
\def\eg{\emph{e.g}\onedot} 
\def\ie{\emph{i.e}\onedot}

\def\etal{\emph{et al}\onedot}
\makeatother

\begin{document}

\title[FOCUS: Frequency-Optimized Conditioning of DiffUSion Models for mitigating catastrophic forgetting during Test-Time Adaptation]{FOCUS: Frequency-Optimized Conditioning of DiffUSion Models for mitigating catastrophic forgetting during Test-Time Adaptation}


\author*[1,2]{\fnm{Gabriel} \sur{Tjio}}\email{ciengabr001@e.ntu.edu.sg}
\author[1]{\fnm{Jie} \sur{Zhang}}\email{{zhang\_jie@cfar.a-star.edu.sg}}
\author[3]{\fnm{Xulei} \sur{Yang}}\email{yang\_xulei@i2r.a-star.edu.sg}
\author[1]{\fnm{Yun} \sur{Xing}}\email{yxing8@ualberta.ca}
\author[1]{\fnm{Nhat} \sur{Chung}}\email{nhatcm25@mp.hcmiu.edu.vn}
\author[4]{\fnm{Xiaofeng} \sur{Cao}}\email{xiaofengcao@tongji.edu.cn}
\author[1]{\fnm{Ivor W.} \sur{Tsang}}\email{lvor\_Tsang@cfar.a-star.edu.sg}
\author[2]{\fnm{Chee Keong} \sur{Kwoh}}\email{ASCKKWOH@ntu.edu.sg}
\author[1]{\fnm{Qing} \sur{Guo}}\email{Guo\_Qing@cfar.a-star.edu.sg}

\affil[1]{\orgdiv{Centre for Frontier AI Research (CFAR)}, \orgname{Agency for Science, Technology and Research (A*STAR)}, \orgaddress{\street{1 Fusionopolis Way, \#16-16 Connexis}, \city{Singapore}, \postcode{138632}, \country{Republic of Singapore}}}
\affil[2]{\orgdiv{College of Computing and Data Science}, \orgname{Nanyang Technological University}, \orgaddress{\street{50 Nanyang Avenue}, \city{Singapore}, \postcode{639798}, \country{Republic of Singapore}}}
\affil[3]{\orgdiv{Institute of Infocomm Research (I2R)}, \orgname{Agency for Science, Technology and Research (A*STAR)}, \orgaddress{\street{1 Fusionopolis Way, \#16-16 Connexis}, \city{Singapore}, \postcode{138632}, \country{Republic of Singapore}}}
\affil*[4]{\orgdiv{School of Computer Science and Technology}, \orgname{Tongji University}, \orgaddress{\street{1238 Gonghexin Road}, \city{Shanghai}, \postcode{200070}, \country{China}}}


\abstract{Test-time adaptation enables models to adapt to evolving domains.
However, balancing the tradeoff between preserving knowledge and adapting to domain shifts remains challenging for model adaptation methods, since adapting to domain shifts can induce forgetting of task-relevant knowledge. 
To address this problem, we propose FOCUS, a novel frequency-based conditioning approach within a diffusion-driven input-adaptation framework.
Utilising learned, spatially adaptive frequency priors, our approach conditions the reverse steps during diffusion-driven denoising to preserve task-relevant semantic information for dense prediction.

FOCUS leverages a trained, lightweight, Y-shaped Frequency Prediction Network (Y-FPN) that disentangles high and low frequency information from noisy images. 
This minimizes the computational costs involved in implementing our approach in a diffusion-driven framework.
We train Y-FPN with FrequencyMix, a novel data augmentation method that perturbs the images across diverse frequency bands, which improves the robustness of our approach to diverse corruptions.

We demonstrate the effectiveness of FOCUS for semantic segmentation and monocular depth estimation across 15 corruption types and three datasets, achieving state-of-the-art averaged performance. 
In addition to improving standalone performance, FOCUS complements existing model adaptation methods since we can derive pseudo labels from FOCUS-denoised images for additional supervision.
Even under limited, intermittent supervision  with the pseudo labels derived from the FOCUS denoised images, we show that FOCUS mitigates catastrophic forgetting for recent model adaptation methods.}

\maketitle

\section{Introduction}\label{sec1}
\begin{figure*} \includegraphics[width=\textwidth]{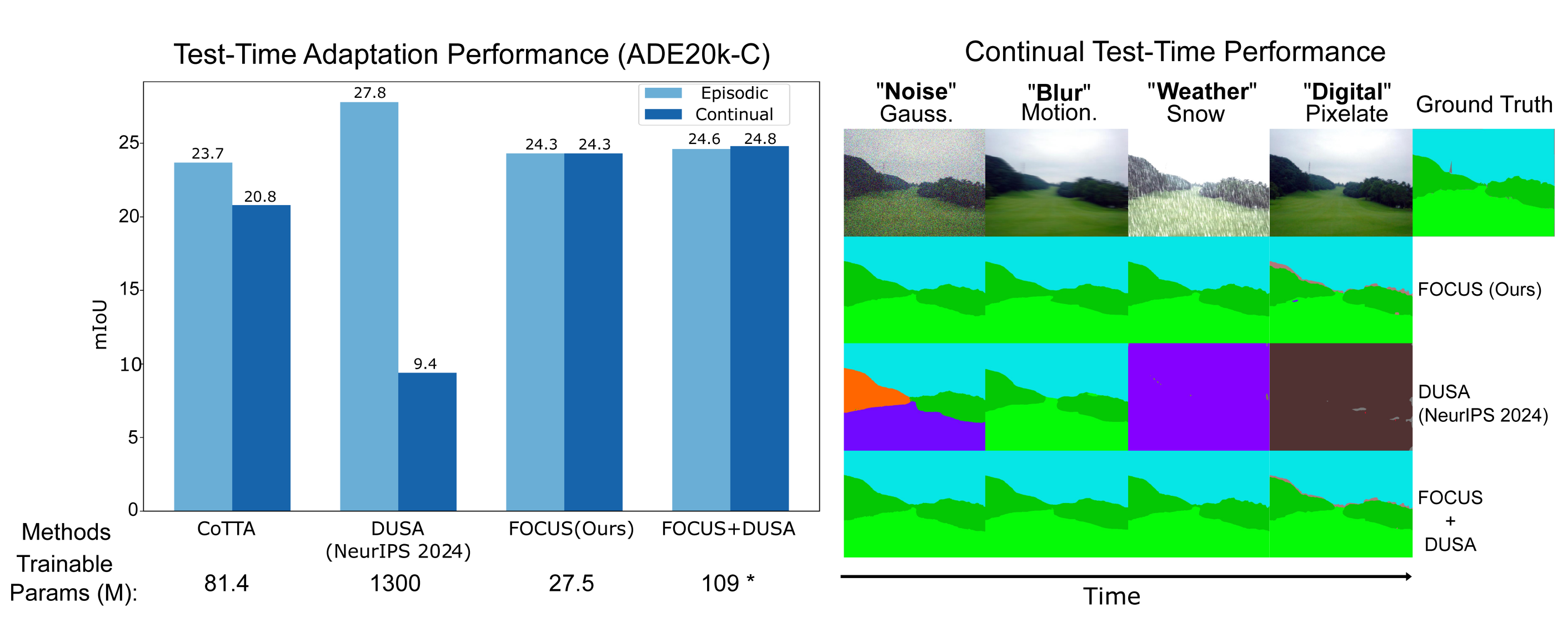}
\captionsetup{width=\textwidth}
\caption{Performance on ADE20k-C. \textbf{Left}:Averaged mIoU across 15 corruptions for \textbf{episodic} (model resets after each corruption) and \textbf{continual} (adapts model across all corruptions without reset) settings.
As a denoising approach, FOCUS is not affected by catastrophic forgetting and recovers degraded performance for DUSA `FOCUS+DUSA'.
\textbf{Right:} Continual test-time qualitative results on representative corruptions from ADE20k-C.
$^*$: DUSA, a diffusion-based method, updates ControlNet, UNet and the task model (Segformer-b5), which raises the number of trainable parameters. 
For `FOCUS$+$DUSA', we only update the task model during adaptation. 
For `FOCUS', we only optimize the parameters for FOCUS before adaptation and do not update the task model.}
\label{fig:fig1_summary}
\end{figure*}
Deep learning models often suffer from significantly reduced performance when deployed in the wild, where test time data distribution diverges from the source domain (training data) distribution  \cite{FilosTMRLG20,Roschewitz2023AutomaticCO}.
To address distribution shifts, model adaptation \cite{Sojka_2024_BMVC,WangSLOD21,Wang_cvpr2022_continual,Li_DUSA_NeurIPS2024}, which improves test-time performance through self-supervised fine-tuning, has emerged as a compelling solution.
However, catastrophic forgetting \cite{Wang_cvpr2022_continual}, where adapting to the test data degrades existing task-relevant knowledge, presents a considerable challenge for model adaptation. 

Catastrophic forgetting occurs when the test data distribution varies over time, and fine tuning the task model with the test data may result in the loss of the original knowledge from the pretrained source model.
While this can be mitigated by preserving the existing knowledge by limiting updates of the task model \cite{Wang_cvpr2022_continual} during test-time adaptation, performance drops noticeably under continual settings (Fig. \ref{fig:fig1_summary}). 
In Fig. \ref{fig:fig1_summary}, catastrophic forgetting erodes performance over time as the model overfits to the test data in each episode, introducing a gap between episodic and continual adaptation performance.

Model adaptation methods generally assume that pseudo label quality is acceptable for model finetuning.
However, this assumption may not hold under extreme, severe conditions.
Input adaptation methods \cite{GaoZLDSW23,SongL2023,HuangYHZCDHWDN23}, which transform the target domain data to match the source domain data, present a possible solution to improve pseudo label quality.
Prior work \cite{GaoZLDSW23} has demonstrated the effectiveness of mapping multiple target domains back to the source data for image classification.
The effectiveness of conditioning diffusion models \cite{Zhang2023_conditionalctrl,ChoiKJGY21,LiuPAZCHSRD23, NicholDRSMMSC22,DhariwalN21, HoS22} in generating user-defined outputs has motivated their use for image denoising \cite{GaoZLDSW23,ChoiKJGY21}.
However, these methods use predefined, image-level priors for conditioning, which may be insufficient for dense prediction.

\subsection{Our Approach}
In this work, we introduce \textbf{learned, spatially adaptive pixel-wise kernels} to enhance the effectiveness of diffusion-driven denoising for dense prediction.
We hypothesise that the semantic information necessary for good performance in dense prediction tasks comprise \textbf{high frequency} (\eg edges) and \textbf{low frequency} (\eg colour) signals, leading us to propose a simple yet effective context-aware frequency prediction network to preserve semantic information.
Since it is non-trivial to manually define robust, generalizable frequency priors, we propose to train a Y-shaped Frequency Prediction Network(Y-FPN) that extracts frequency priors from noisy images.
We then use these priors to condition diffusion-driven denoising for dense prediction (segmentation and depth estimation).
\textbf{To the best of our knowledge, this is the first work to condition diffusion denoising via learned pixel-level frequency priors for dense prediction tasks}.
We list our contributions below:
\begin{itemize}
    \item Our novel workflow, \textbf{F}requency-\textbf{O}ptimized \textbf{C}onditioning in Diff\textbf{US}ion (FOCUS) introduces a novel approach that mitigates catastrophic forgetting by using spatially adaptive frequency priors to condition a pretrained diffusion model for diffusion-driven denoising.
    We extract the learned frequency priors with a lightweight Y-shaped Frequency Prediction Network (Y-FPN). 
    To train Y-FPN, we introduce FrequencyMix to diversify training data.  
    \item \textbf{Performance:} We evaluate FOCUS on severe, diverse corruptions ($n$=15) from the ImageNet-C dataset and show that FOCUS outperforms best existing work on averaged performance for \textbf{segmentation}- ADE20k-C: (DUSA \cite{Li_DUSA_NeurIPS2024}:+14.9$\%$ mIoU, DDA\cite{GaoZLDSW23}:1.7 $\%$ mIoU); Cityscapes-C:(CoTTA \cite{Wang_cvpr2022_continual}: +3.7$\%$ mIoU) and \textbf{depth estimation} -NYU2k-C: (DDA \cite{GaoZLDSW23}:-0.32 relative error).
 \item  \textbf{Complementarity:} As a task agnostic, input adaptation method, FOCUS can be seamlessly combined with model adaptation techniques.
FOCUS yields synergistic gains for both \textbf{segmentation} and \textbf{depth estimation}, validating its utility as a flexible, modular addition to CTTA pipelines. Importantly, FOCUS benefits model adaptation even with \textbf{intermittent supervision} for semantic segmentation (\eg (ADE20k-C) `FOCUS+TENT ($k$=25)':+0.5\% mIoU, where the pseudo labels from the denoised images are applied \textbf{once every 25 iterations}) and depth estimation: `FOCUS+CoTTA ($k$=50)' :-0.06 relative error), highlighting the effectiveness of FOCUS. 
\end{itemize}

\section{Related Work}
\subsection{Test-time domain adaptation}
Test-time adaptation can be categorized into \textit{model adaptation} and \textit{input adaptation}.
Model adaptation methods finetune the model during inference
Since the ground truth labels are unavailable at test time, model adaptation methods rely on self supervision approaches such as entropy minimization \cite{WangSLOD21}, pseudo labeling \cite{Kundu2021genadapt, Wang_cvpr2022_continual},  and feature distribution alignment \cite{liang2020shot}.
Since the pseudo labels are noisy, much effort has been devoted to improving pseudo label reliability using methods such as consistency regularization \cite{Wang_cvpr2022_continual}, which encourages similar model output from augmented instances of the same input image and by pseudo label filtering \cite{rectify_pseudolabels}.
Although numerous prior studies \cite{HuSGLCCZZ21, WangSLOD21, SivaprasadF21, Wang_cvpr2022_continual, Li_DUSA_NeurIPS2024, ShinTZSLGKY22,Gong_2024_BMVC} study TTA for dense prediction tasks, the majority of existing work finetune on the test data. 

Input adaptation methods transform the styles of the target domain data to match the styles of the original training data (source domain images) while preserving semantic information of the target domain images.
Earlier work \cite{Hoffman_2018_cycada,kim2020_textureinvariant} rely on Generative Adversarial Networks (GANs), which perform a one-to-one transformation from the target domain images to the source domain images. 
Unlike GAN-based approaches \cite{Hoffman_2018_cycada}, diffusion models offer a greater degree of fine-grained control over the generated outputs \cite{NieGHXVA22, GaoZLDSW23, SongL2023, HuangYHZCDHWDN23} and map the target domains to the source domain data (many-to-one transformation). 
Kim \etal \cite{kim2020_textureinvariant} observed that arbitrary style transfer-based approaches introduce image artifacts in the generated images which can hinder adaptation performance. 
\textbf{While our work updates the input (input adaptation) via image denoising, we also show that our work can be integrated with existing model adaptation methods to improve performance.} 

\subsection{Diffusion models}
Diffusion modelling has recently gained widespread prominence and demonstrates strong generative capacity \cite{HoJA20, Song0011SKKEP21, SongE19, Song0011E20, Sohl-DicksteinW15, BlattmannRLD0FK23, NicholD21}.
The underlying concept for diffusion (\ie denoising diffusion probabilistic model (DDPM) \cite{HoJA20}) can be described as follows -  the `forward process' iteratively adds noise to the data, and the `reverse process' leverages a trained network to progressively recover the original data through a series of steps.

Diffusion models show a high degree of user controllability via conditioning on visual cues \cite{ChoiKJGY21}, natural language \cite{LiuPAZCHSRD23, NicholDRSMMSC22}, and class labels \cite{DhariwalN21, HoS22}. 
Recent work \cite{GaoZLDSW23,oh2024efficient,Tsai_2024_CVPR} demonstrated the effectiveness of diffusion models in projecting non-Gaussian corrupted data to the source distribution.
In particular, Gao \etal \cite{GaoZLDSW23} show the effectiveness of a predefined image-level low-pass filter for robust test-time input adaptation of image classification tasks.
Gao \etal \cite{GaoZLDSW23} used predefined image-level filters to preserve the image-level semantic information of the corrupted images during denoising.
Unlike their work, we propose to learn the high and low frequency priors for each pixel to condition diffusion-driven denoising.

\subsection{ Frequency-aware processing}
There is substantial interest in applying frequency modelling to guide the development of robust, generalizable vision systems \cite{2023iccv_PASTA,ChoiKJGY21,GaoZLDSW23,Xie0002LZ0OL23,XuZFWWZ23}.
It was previously observed that classification models are biased towards colour and texture information \cite{geirhos2018imagenettrained}, and that vision models make predictions based on a combination of low and high-frequency information \cite{Yin_fourier_robust_NIPS2019,Wang_2023_ICCV}. 
Additionally, prior studies\cite{2023iccv_PASTA,amp_phase_visual_imp} show that the frequency spectrum presents a representational space to separate and extract the underlying patterns and semantic information from a image.
This is useful for data augmentation, especially since the goal is to increase data diversity without changing the semantic meaning of the images.
Earlier work \cite{2023iccv_PASTA} preferentially perturb the high-frequency amplitude spectrum compared to the low-frequency amplitude spectrum of the training data to improve adaptation performance to real-world data (Synthetic training data \textrightarrow Real-world semantic segmentation datasets).
However, since our goal to improve generalizability to a broad spectrum of noise and image corruptions, we train our model using a composite augmented image generated from our novel approach, FrequencyMix, which combines the differently augmented images.

In addition to training robust models, frequency-aware processing can also be used to condition generative priors.
Notably, Choi \etal \cite{ChoiKJGY21} and Gao \etal \cite{GaoZLDSW23} demonstrated the effectiveness of conditioning the diffusion model with predefined image-level low-pass filters for user-defined image generation. 
We extend their work and introduce learned, spatially adaptive low and high-frequency priors to condition a pretrained diffusion model to mitigate catastrophic forgetting during test-time adaptation.

\section{Methods}
\begin{figure*}  \includegraphics[width=\textwidth]{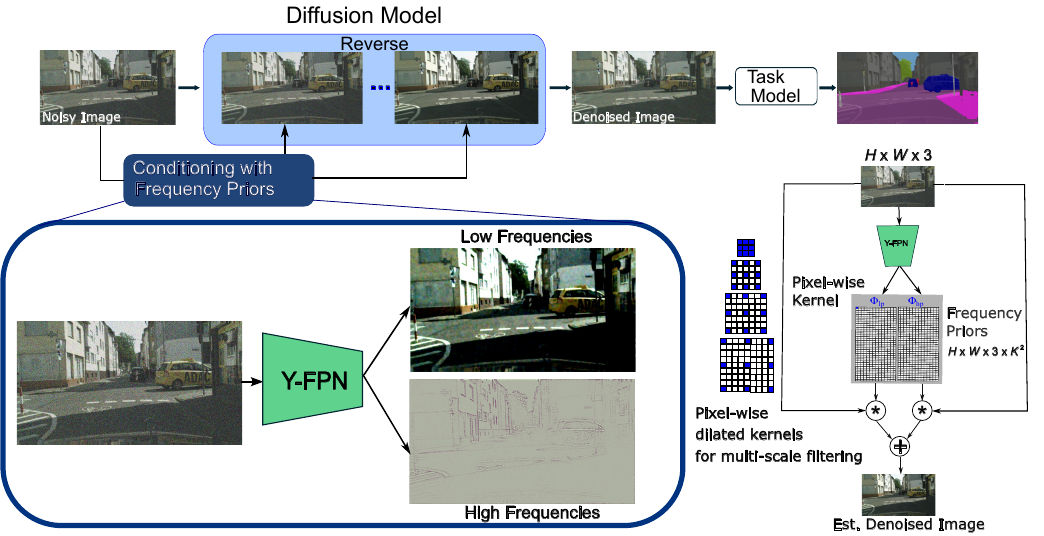}
\captionsetup{width=\textwidth}
\caption{\textbf{FOCUS Workflow.} We condition the reverse denoising steps with our spatially adaptive frequency priors. The Y-shaped Frequency Prediction Network (Y-FPN) outputs the  frequency priors ($\Phi_{\text{lp}}$ and $\Phi_{\text{hp}}$) of size $K\times K \times 3$ for each pixel of the input  noisy RGB image with dimensions $H\times W \times 3$ pixels.}
\label{fig:fig3_workflow}
\end{figure*}

In this section, we introduce FOCUS (Fig. \ref{fig:fig3_workflow}), a frequency-conditioned diffusion-driven input adaptation method. 
We first introduce the theoretical background in Section \ref{subsec:frequency_priors_diffusion_theory} for conditioning diffusion-driven denoising.
In Section \ref{subsec:FPN visual filtering}, we introduce our lightweight (27.5M parameters) Y-shaped Frequency Prediction Network (Y-FPN) which generates spatially adaptive frequency priors. 
We also show the loss functions used to train Y-FPN (Section \ref{subsec:YFPN_loss_fn}) and the test-time loss functions (Section \ref{subsec:TTA_loss}) used for additional supervision from the FOCUS denoised images.

\subsection{Leveraging Frequency Priors to Condition Denoising via Diffusion}
\label{subsec:frequency_priors_diffusion_theory}
Given an input image $\boldsymbol{x}_0$ randomly sampled from a target distribution $q(\boldsymbol{x}_0)$, we leverage a ImageNet pretrained diffusion model \cite{DhariwalN21} to map $\boldsymbol{x}_0$ to the source domain.

Diffusion comprises a forward process, which iteratively adds Gaussian noise to $\bm{x}_0$ over $N$ steps ($N$=1000), and a reverse process, which estimates the input image $\boldsymbol{x}_{t}^g$ for each step $t$.
$g$ indicates guidance (\eg via frequency priors). 

For step $t-1$, we sample the estimate $\hat{\boldsymbol{x}}_{t-1}^g$ from the conditional distribution, \ie, $\hat{\boldsymbol{x}}_{t-1}^g\sim p_{\theta}({\boldsymbol{x}}_{t-1}^g | {\boldsymbol{x}}_{t}^g)$, where $p_{\theta}$ is the parameterized learned process that approximates the reverse of the forward noising steps.

To preserve essential semantic information during diffusion-driven denoising for image classification, Gao \etal \cite{GaoZLDSW23} introduce an image-level, low-frequency preserving constraint to the reverse process.
For the time step $t-1$,  Gao \etal  estimate the denoised image $\hat{\boldsymbol{x}}_{0}^g$ by
\begin{align}\label{eq:reverse-purify}
		\hat{\boldsymbol{x}}^g_0 = \sqrt{\frac{1}{\overline{\alpha}_t}}\boldsymbol{x}^g_t - \epsilon_\theta(\boldsymbol{x}_t^g, t)\sqrt{\frac{1}{\overline{\alpha}_t} - 1}.
\end{align}
${\alpha}_t:=1-\beta_{t}$,$\overline{\alpha}_t:=\prod_{t=1}^{N}\alpha_{t}$, where $\beta_{t}$ refers to the noise added at each step $t$ during the forward diffusion process.

Finally, they update the estimate $\boldsymbol{x}^g_{t-1}$ from the unconditional proposal $\hat{\boldsymbol{x}}_{t-1}^g$ by conditioning with the predefined low frequency prior $\Phi_\text{lp'}$,
\begin{align}\label{eq:reverse-process-guidance}
\boldsymbol{x}^g_{t-1} = \hat{\boldsymbol{x}}_{t-1}^{g} - w \nabla_{\boldsymbol{x}_t}\left\|\Phi_\text{lp'}\left(\boldsymbol{x}_{0}\right) -\Phi_\text{lp'}\left(\hat{\boldsymbol{x}}^g_0\right)\right\|_2,
\end{align}
where $\Phi_\text{lp'}(\cdot)$ is the image-level, predefined low-pass filter, and the scale weight $w$ controls the step size of the guidance.
They derive the low-pass filtering effect by downsampling the image followed by upsampling back to the original image dimensions.
They then iteratively update the estimate of the denoised image $\boldsymbol{x}^g_{t-1}$ until $t=0$.

We derive the denoised image $\boldsymbol{x}^g_{t-1}$ from the unconditional proposal $\hat{\boldsymbol{x}}_{t-1}^g$,
\begin{equation}
\begin{split}
\label{eq:reverse-process-guidance-freqadded}
\boldsymbol{x}^g_{t-1} = & \hat{\boldsymbol{x}}_{t-1}^{g} - w \nabla_{\boldsymbol{x}_t}(\left\|\Phi_\text{lp}\left(\boldsymbol{x}_{0}\right) -\Phi_\text{lp}\left(\hat{\boldsymbol{x}}^g_0\right)\right\|_2 \\ &+\left\|\Phi_\text{hp}\left(\boldsymbol{x}_{0}\right) -\Phi_\text{hp}\left(\hat{\boldsymbol{x}}^g_0\right)\right\|_2),
\end{split}
\end{equation}
where $\Phi_\text{lp}(\cdot),\Phi_\text{hp}(\cdot)$ are our learned, spatially adaptive low and high frequency priors respectively, and the scale weight $w$ controls the step size of the guidance. 
We iteratively update the estimate of the denoised image  $\boldsymbol{x}^g_{t-1}$. 
The estimate of the denoised image is then passed to the task model for segmentation/depth estimation.

\subsection{Y-shaped Frequency Prediction Network for Visual Filtering}
\label{subsec:FPN visual filtering}
In this section, we present our frequency-based filtering method for robust test-time input adaptation.
We process a corrupted image $\mathbf{I}^{c} \in \mathbb{R}^{H\times W}$ for pixel-wise filtering, 
\begin{equation}
\mathbf{I}^{f}=\sum_{i=1}^{l=2} \mathbf{K}_{i} \odot \mathbf{I}^{c}, 
\label{eqn:imagecorruptions}
\end{equation}
where $\mathbf{I}^{f}\in \mathbb{R}^{H\times W}$ is the estimate of the denoised image and $\odot$ denotes the pixel-wise filtering operation.
$\mathbf{K}_{i}\in \mathbb{R}^{H\times W \times K^{2}}$ denotes the kernels of size $K\times K$ ($K=3$).
We jointly apply $\mathbf{K}_{i=1,2}$ to the corrupted image $\mathbf{I}^{c}$ to generate $\mathbf{I}^{f}$.

Encouraged by the successful application of Kernel Prediction Networks (KPN) across several tasks \cite{Mil_burstdenoise_2018, limisf_2022, Guo_efficientderain_2021}, we leverage KPNs to learn the spatially adaptive kernels for conditioning diffusion models.
We propose to estimate the kernels $\mathbf{K}_{i=1,2}$ from a noisy image $\mathbf{I}^{c}$,
\begin{equation}
\mathbf{K}_{i=1,2}= \text{Y-FPN} 
(\mathbf{I}^{c}),
\label{eqn:kpns}
\end{equation}
where Y-FPN denotes the Y-shaped Frequency Prediction Network and shares a similar architecture with UNet \cite{unet}.
Unlike previous work \cite{Mil_burstdenoise_2018,limisf_2022, Guo_efficientderain_2021}, which learn a single kernel for each task, Y-FPN predicts 2 separate kernels $\mathbf{K}_{i=1}$ and $\mathbf{K}_{i=2}$ per pixel for a given image. 

For a given pixel $p$ in the corrupted image $\mathbf{I}^{c}$, we derive the low frequency priors $\mathrm{\Phi}_{lp}$  from $\mathbf{K}_{i=1}$ by applying the method \cite{zou2020delving}, which normalizes
$\mathbf{K}_{i=1}$ with a spatial softmax:
\begin{equation}
\mathbf{K}_{i=1}[u,v] = \frac{\exp(\mathbf{K}_{i=1}[u,v])}{\sum_{u',v'} \exp(\mathbf{K}_{i=1}[u',v'])},
\label{eqn:spatial_softmax}
\end{equation}
where $(u,v)$ indexes the spatial location in the $K\times K$ kernel. 
We then compute the low-frequency prior for pixel $p$
\begin{equation}
\Phi_{\mathrm{lp}}(\mathbf{I}^{c})[p] = \sum_{u,v} \mathbf{K}_{i=1}[u,v].  \mathbf{I}^{c}[p].
\end{equation}
We also normalize $\mathbf{K}_{i=2}$ with a spatial softmax (Eqn \ref{eqn:spatial_softmax})  and compute the following equation to obtain the high frequency prior for pixel $p$
\begin{equation}
\Phi_{\mathrm{hp}}[p](\mathbf{I}^{c}) = \mathbf{I}^{c} - \sum_{u,v} \mathbf{K}_{i=2}[u,v] . \mathbf{I}^{c}[p].
\end{equation}

To train Y-FPN, we apply the commonly used loss functions for image restoration \ie{} $L_{1}$,  SSIM (Structural Similarity)   and a Frequency Reconstruction Loss \cite{Kim_BMVC2021_Freq}.

\subsection{Loss Functions for Training Y-FPN}
\label{subsec:YFPN_loss_fn}
We apply the commonly used loss functions for image restoration \ie{} $L_{1}$,  SSIM (Structural Similarity)  to train our network.
We also include a Frequency Reconstruction Loss \cite{Kim_BMVC2021_Freq} $\mathcal{L}_{Freq}$.   
We first apply the Fourier Transform to the filtered image $\mathbf{I}^{f}$ and the original clean image $\mathbf{I}$.
We then compute the difference between Fourier transformed images and normalize the result with a logarithmic function
\begin{equation}
\mathcal{L}_{Freq}(\mathbf{I}^{f},\mathbf{I})=\log(1 + \frac{1}{HW} \| \mathcal{F}(\mathbf{I}) - \mathcal{F}(\mathbf{I}^{f}) \|) ,
\label{eqn:freqloss}
\end{equation}
where $H, W$ are the spatial dimensions of the image in frequency space.
Bringing it all together,  we have the following overall training loss for Y-FPN,
\begin{equation}
\mathcal{L}(\mathbf{I}^{f},\mathbf{I})=\|\mathbf{I}^{f}-\mathbf{I} \|_{1} - \lambda_{1} \text{SSIM}(\mathbf{I}^{f},\mathbf{I}) + \lambda_{2}\mathcal{L}_{Freq}(\mathbf{I}^{f},\mathbf{I}).
\label{eqn:kpnloss}
\end{equation}
where $\lambda_{1},\lambda_{2}$ are hyperparameters.
We fix $\lambda_{1},\lambda_{2}$=0.2 for all experiments.

\subsection{Loss functions for Test-Time Adaptation}
\label{subsec:TTA_loss}
Here, we provide a brief description of the loss functions during test-time adaptation.
For semantic segmentation, we provide additional supervision from the FOCUS denoised images via the standard cross entropy loss,
\begin{equation}
\mathcal{L}_{FOCUS}(G,\boldsymbol{X},\boldsymbol{Y^{'}}) = \sum_{i=1}^{H \times W}\sum_{c=1}^{C} -Y^{'}_{ic}log(G(\boldsymbol{X})),
\label{eqn:cross_ent_eqn}
\end{equation}
where $G(\boldsymbol{X})$ refers to the predicted probability of class $c$ for the \textit{i}th pixel for the input image $\boldsymbol{X}$. $Y^{'}_{ic}$ is the predicted label from the task network $G$ on the FOCUS denoised images for class $c$ on the \textit{i}th pixel, where $Y^{'}_{ic}=1$ if the pixel belongs to the class \textit{c} and $Y^{'}_{ic}=0$ if otherwise.

In preliminary experiments for depth estimation on NYU2k-C, we used the standard SILog loss \cite{Eigen_NIPS2014_depth} to update the baseline model adaptation methods.
However, we observed that the model adaptation methods converged to degenerate solutions (e.g., predicting constant depth), which could be caused by unstable gradients in the presence of strong corruptions.

To enable a fair comparison and ensure convergence, we use the L1 loss, which yielded more stable results.
We provide a description below on the loss functions used for depth estimation.
Given an input $\boldsymbol{x}$ and its horizontally flipped image $\boldsymbol{x}_{flip}$, we compute the depth using the depth estimation model $D$.
\begin{equation}
\boldsymbol{p} = D(\boldsymbol{x}),
\label{eqn:depth_pred}
\end{equation}
\begin{equation}
\boldsymbol{p_{flip}} = D(\boldsymbol{x}_{flip}),
\label{eqn:depth_predflip}
\end{equation}
We derive the continuous-valued confidence mask $s$ as follows
\begin{equation}
s = \frac{1}{|\boldsymbol{p_{flip}}-\boldsymbol{p}|+10^{-6}}.\label{eqn:confidence_mask}
\end{equation}
We then optimize the model adaptation methods using the following consistency loss
\begin{equation}
\mathcal{L}(D,\boldsymbol{x},\boldsymbol{x}_{flip}) = |\boldsymbol{p_{flip}}-\boldsymbol{p}| + s|\boldsymbol{p}-D(\boldsymbol{x})|.
\label{eqn:depth_consistency_loss}
\end{equation}

For FOCUS-driven supervision, we include the FOCUS denoised images $\boldsymbol{x}_{FOCUS}$,
\begin{equation}
\mathcal{L}_{FOCUS}(D,\boldsymbol{x},\boldsymbol{x}_{FOCUS}) =   s|\boldsymbol{p}-D(\boldsymbol{x}_{FOCUS})|.
\label{eqn:seg_consistency_loss}
\end{equation}

\subsection{FrequencyMix Training}
\label{subsec:training}
In this section, we introduce our approach to generate sufficiently challenging data for training the Y-shaped Frequency Prediction Network (Y-FPN).
Since Y-FPN must be robust to noise across a broad range of frequencies, we apply \textbf{frequency-dependent perturbations} and  combine these frequency-dependent perturbations via  \textbf{FrequencyMix} to increase the training data diversity.

\subsubsection{Frequency Dependent Perturbations}
\label{subsubsec:freq_dep_perturb}
Inspired by the success of previous work \cite{2023iccv_PASTA} that improved model generalizability by perturbing the amplitude spectrum of the high spatial frequency region, we adapted their approach to include additional  perturbations.

For a single-channel image $\bm{x} \in \mathbb{R}^{H\times W}$, the Fourier transform $\mathcal{F}(\boldsymbol{x})$ is described as
\begin{equation}
\begin{split}
&\mathcal{F}(\boldsymbol{x})[m,n]=\\&\sum_{h=0}^{H-1}\sum_{w=0}^{W-1}\boldsymbol{x}[h,w]\exp \left({-2\pi i(\frac{h}{H}m+\frac{w}{W}n)}\right),
\label{eqn:fouriertxf}
\end{split}
\end{equation}
where $i^{2}=-1$ and $m,n$ refer to the spatial frequencies in Fourier space.

We can then derive the amplitude spectrum $\mathcal{A}(\bm{x})$ and the phase spectrum $\mathcal{P}(\boldsymbol{x})$ as follows
\begin{equation}
    \mathcal{A}(\bm{x})=\sqrt{(\Re(\mathcal{F}(\bm{x})[m,n])^2+\Im(\mathcal{F}(\bm{x})[m,n])^2)},
    \label{eqn:amp_spec_def}
\end{equation}
\begin{equation}
    \mathcal{P}(\boldsymbol{x})= arctan\left(\frac{\Im(\mathcal{F}(\bm{x})[m,n]}{\Re(\mathcal{F}(\bm{x})[m,n]}\right)    ,\label{eqn:phase_spec_def}
\end{equation}
where $\Re(\mathcal{F}(\bm{x})[m,n])$ and $\Im(\mathcal{F}(\bm{x})[m,n])$ denote the real and imaginary components of the Fourier Transform respectively.
Since we are working with RGB images, we compute the Fourier Transform for each channel independently.

To perturb the clean images, we introduce $\epsilon_{high}[m,n]$, $\epsilon_{low}[m,n]$, $\epsilon_{rand}[m,n]$ and $\epsilon_{uni}$.
We use the spatially dependent functions to control the perturbation extent.
\begin{equation}
\epsilon_{high}[m,n] \sim \mathcal{N}\left(1,\left(2\alpha\sqrt{\frac{m^{2}+n^{2}}{H^{2}+W^{2}}}+\beta\right)^2\right),
\label{eqn:epilson_high}
 \end{equation}
 \begin{equation}
 \epsilon_{low}[m,n]\sim\beta\mathcal{N}\left(1,2 e^{-\alpha(m^{2}+n^{2})}\right),
\label{eqn:epilson_low}
 \end{equation}
\small
 \begin{equation}
 \begin{split}
&\epsilon_{rand}[m,n] \\&\sim\mathcal{N}\left(1,\alpha\frac{(H-m+y_{pk})^{2}+(W-n+x_{pk})^{2}}{H^{2}+{y_{pk}}^2+W^{2}+{x_{pk}}^2}\right)
\end{split}
\label{eqn:epilson_rand}
 \end{equation}
\begin{equation}
\epsilon_{uni}=\alpha \beta,
\label{eqn:epilson_uni}
 \end{equation}
where  severity $\alpha \sim (3,4,5)$, $x_{pk} \sim Uniform(0,W/4)$, $y_{pk} \sim Uniform(0,H/4)$, $\beta=0.3$, \\ $m \in [-H/2,H/2]$, $n \in [-W/2,W/2]$.
We zero-centre the amplitude spectrum $\mathcal{A}(\boldsymbol{x})$ before applying the perturbation function $g(.)$
\begin{equation}
\begin{aligned}\hat{\mathcal{A}}(\boldsymbol{x})&=g(\mathcal{A}(\boldsymbol{x}),\epsilon )[m,n] \\&= (1+\epsilon \mathcal{A}(\boldsymbol{x})),
\label{eqn:amplitude perturbation}
   \end{aligned}
\end{equation}
where  $\epsilon
 \in [\epsilon_{rand},\epsilon_{high},\epsilon_{low},\epsilon_{uni}]$.
Finally, we obtain the augmented image via 2D-FFT by recombining the perturbed amplitude spectrum $\hat{\mathcal{A}}(\boldsymbol{x})$ with the original phase spectrum $\mathcal{P}(\boldsymbol{x})$.

\begin{figure}[htp!]  \includegraphics[width=\columnwidth]{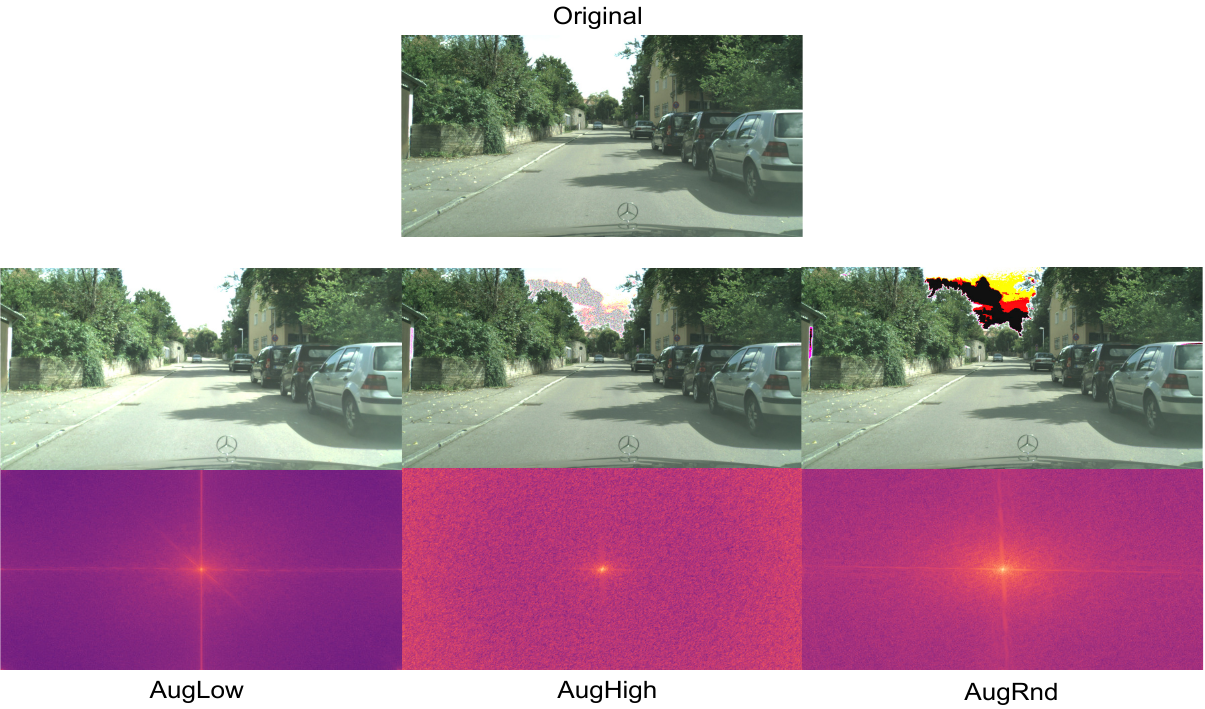}
\captionsetup{width=\columnwidth}
\caption{Examples of amplitude perturbed training images `AugLow', `AugHigh' and `AugRnd'.
Top Row: Augmented Images `AugLow', `AugHigh' and `AugRnd'.
Bottom Row: FFT of the difference images (Augmented Image-Original).}
\label{fig:fig_freqmix_examples}
\end{figure}

\subsubsection{FrequencyMix}
\label{subsubsec:Freqmix}
 
 \begin{figure}
 \centering
 \includegraphics[width=\columnwidth]{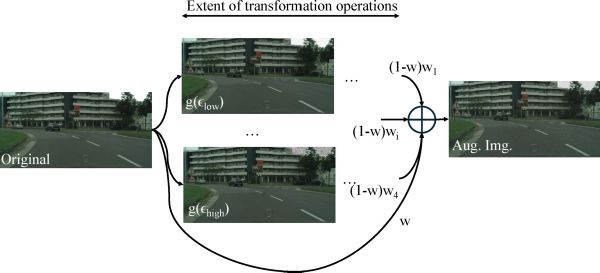}
 \captionsetup{width=\columnwidth}
 \caption{\textbf{Illustration of FrequencyMix}. We combine randomly selected augmentations and re-weight the augmented images to generate a composite augmented image. The weights $w_{i}$ are sampled from a Dirichlet distribution, and the blending weight $w$ is sampled from a Beta distribution.}
 \label{fig:freqmix} 
 \end{figure}
While we have addressed the difficulty of perturbing images as a function of spatial frequency, we are still faced with the challenge of generating sufficiently diverse data for training Y-FPN.
Previous work demonstrated the effectiveness of data augmentation by randomly sampling and combining data transformation operations for robust image classification   \cite{hend2020augmix} and image de-raining \cite{Guo_efficientderain_2021}.
Similar to Augmix \cite{hend2020augmix}, we randomly sample and combine the frequency-dependent perturbations: low-frequency, high-frequency and randpeak (randomized frequency peaks).
\begin{algorithm*}
\caption{FrequencyMix and Y-FPN training}
 \begin{algorithmic}[1]
       \INPUT  Clean Images $\mathcal{I}$, Y-like Frequency Prediction Network Y-FPN(.),\newline Loss function $\mathcal{L}$.,   Augmentations $Aug=[\text{Uni. Noise},  g(\epsilon_{low}), g(\epsilon_{high}), g(\epsilon_{uni}), g(\epsilon_{rand})$]
   \OUTPUT Trained Y-like Frequency Prediction Network Y-FPN(.) \newline   \Procedure{\textbf{Function} FrequencyMix}{(clean Image $\mathbf{I}):$}
      \myState{Sample mixing weights $(w_{1},w_{2},w_{3},w_{4})\sim Dirichlet;$}
     \myState{Initialize an empty image $\mathbf{I}_{mix}$;}
 \FOR{$i=1$, ..., $4$}
    \myState{Sample augmentations $(a_{1},a_{2},a_{3},a_{4}) \sim Aug $;}
   \myState{Combine augmentations via $a_{12}=a_{2}a_{1}$,  $a_{123}=a_{3}a_{2}a_{1}$ and  $a_{1234}=a_{4}a_{3}a_{2}a_{1}$;
   }
    \myState{Sample operation $ o \sim (a_{1},a_{12},a_{123},a_{1234})$; }
\myState{$\mathbf{I}_{mix}+=w_{i}o(\mathbf{I})$;}
   \ENDFOR
   \myState{Sample a blending weight $w\sim Beta$;}
   \myState{\textbf{return} $\mathbf{I}_{noisy}= w\mathbf{I} + (1-w)\mathbf{I}_{mix}$}
   \EndProcedure
   \FOR{$i=0$, ..., $iter_{num}$}
       \myState{Sample a clean Image $\mathbf{I} \sim \mathcal{I}$;}
   \myState{Generate noisy image with FrequencyMix $\mathbf{I}_{noisy}\leftarrow$ FrequencyMix($\mathbf{I}$); }
   \myState{Generate estimate of clean image with Y-FPN $\hat{\mathbf{I}} \leftarrow \text{Y-FPN}(\mathbf{I}_{noisy})$;}
   \myState{Calculate Loss $\mathcal{L}$ from eqn \ref{eqn:kpnloss} and backpropagate ;}
   \myState{Update Y-FPN(.) parameters. }
\ENDFOR
 \end{algorithmic}
  \label{alg:Freqmix_KPN}
 \end{algorithm*}

\section{Experimental Details}
In this section, we provide an overview of the datasets (Section \ref{subsec:datasets}), baselines (Section \ref{subsec:baselines}) and implementation details (Section \ref{subsec:Imp_details}). 

 \subsection{Datasets}
 \label{subsec:datasets}
 Here, we describe the datasets used in this work.
\begin{itemize}
\item ImageNet-C \cite{HendrycksD19} We apply the corruptions drawn from the categories (\ie{} \textit{Noise}, \textit{Blur}, \textit{Digital}, \textit{Weather}) in ImageNet-C \cite{HendrycksD19} at the highest severity (level=5) to the evaluation images.
    \item ADE20k \cite{zhou2017scene,zhou2019semantic} is a densely annotated semantic segmentation dataset of indoor and outdoor images.
We use 25,574 images for training Y-FPN and 2,000 images for evaluation.
Since there is randomness involved during the corrupted image generation, we use the publicly available ADE20k-C validation set from Li \etal \cite{Li_DUSA_NeurIPS2024} to preserve reproducibility.
    \item Cityscapes \cite{Cordts2016Cityscapes} is a real-world driving semantic segmentation dataset with densely annotated images of resolution 2048 $\times$ 1024 pixels.
     We train Y-FPN with  the default split of 2,975 images.
    We apply the corruptions from ImageNet-C on the evaluation split of 500 images.
    \item NYU2k \cite{Silberman_NYU2k} is a densely annotated dataset of indoor scenes for depth estimation.
    We train Y-FPN with the standard split of 50,688 images and evaluate performance on the test split of 654 images at the original resolution of 640 $\times$ 480 pixels after applying the corruptions from ImageNet-C.
\end{itemize}

\subsection{Baselines}
\label{subsec:baselines}

\begin{itemize}
\item TENT \cite{WangSLOD21} is a keystone paper for test-time adaptation, having demonstrated effectiveness for image classification and semantic segmentation.
It updates the affine parameters within the batch normalization layers via entropy minimization.
Since entropy is not well-defined for depth estimation, we apply a consistency loss between original and augmented predictions.
    \item CoTTA \cite{Wang_cvpr2022_continual} is a model adaptation method that utilizes a exponential mean weighting average model as a teacher to generate reliable pseudo labels for test time adaptation.
    CoTTA addresses catastrophic forgetting by stochastically updating the weights from the teacher model with the weights from the pretrained source model.
    \item DDA  \cite{GaoZLDSW23} utilizes a predefined image-level low-pass filter to condition a pretrained ImageNet diffusion model for denoising of corrupted images in pixel-space.
    Unlike model adaptation methods that are sensitive to batch sizes and learning rates, DDA is invariant to these test-time hyperparameters.
    \item DUSA \cite{Li_DUSA_NeurIPS2024} is a recent state-of-the-art work utilising a latent space diffusion model for test-time adaptation.
    Leveraging a ControlNet trained on ADE20k, they use the prediction logits to condition the diffusion model.
    They then estimate the noise at a single timestep (N=100) and optimize the task model and the diffusion model by minimizing the error between the estimated noise and the actual noise. 
    We use their publicly available code for comparison.
\end{itemize}

\subsection{Implementation Details}
\label{subsec:Imp_details}
We use the publicly available pretrained models (Segformer-b5 \cite{xie2021segformer} for semantic segmentation, and DenseDepth \cite{Alhashim2018} for depth estimation).
For DDA, we use the same number of timesteps and scaling weight as FOCUS.
For Y-FPN training, we use learning rate $2\times10^{-4}$, Adam optimizer with $\beta_{1}=0.5,\beta_{2}=0.999$.
We describe FrequencyMix and Y-FPN training in Algorithm \ref{alg:Freqmix_KPN} and show the FOCUS workflow in Algorithm \ref{alg:supp_FDD_alg}.

\noindent \textbf{Semantic Segmentation}
For semantic segmentation tasks, we use the implementation for TENT, CoTTA and DUSA from Li \etal \cite{Li_DUSA_NeurIPS2024}.
We perform model adaptation (TENT, CoTTA, DUSA) with a batch size of 1, Adam optimizer with a learning rate of $6\times10^{-5}/8$.
\noindent For Cityscapes, we use the hyperparameters $\lambda_{1}=0.2,\lambda_{2}=0.2$ to train Y-FPN.
During diffusion-driven denoising, we use scale weights $w=2$ and the number of timesteps $nsteps=15$.

\noindent For ADE20k, we train Y-FPN with $\lambda_{1}=0.2,\lambda_{2}=0.2$.
During diffusion-driven denoising, we use scale weights $w=2$ and the number of timesteps, $nsteps=25$.

\noindent  \textbf{Depth Estimation}
For the model adaptation methods (TENT, CoTTA), we use a batch size of 1, Adam optimizer.
We evaluate performance at learning rate $1\times10^{-4}$ and $1\times10^{-5}$.
we train Y-FPN with $\lambda_{1}=0.2,\lambda_{2}=0.2$.
During diffusion-driven denoising, we use scale weights $w=3$, and the number of timesteps, $nsteps=25$.

\begin{algorithm}
\caption{\textbf{F}requency \textbf{O}ptimized \textbf{C}onditioning for Diff\textbf{US}ion}
\small
 \begin{algorithmic}[1]
  \INPUT  Noisy Image $\boldsymbol{I_{noisy}}$, Trained Y-like Frequency Prediction Network Y-FPN(.), Pretrained (ImageNet) Denoising Probabilistic Diffusion Model, Number of Timesteps $n$ \OUTPUT Denoised image $\boldsymbol{x_{0}^{g}}$
\myState{ Obtain estimate of denoised image  $\boldsymbol{x_0}\leftarrow \text{Y-FPN} (\boldsymbol{I_{noisy}})$ };
\myState{Sample $\boldsymbol{x}_N \sim q(\boldsymbol{x}_N|\boldsymbol{x}_0)$}
 \FOR{$t=1000$...$1000-n$}
\myState{$\hat{\boldsymbol{x}}_{t-1}^g\sim p_{\theta}({\boldsymbol{x}}_{t-1}^g | {\boldsymbol{x}}_{t}^g)$;}
\myState{
$\boldsymbol{x}^g_{t-1}\leftarrow \hat{\boldsymbol{x}}_{t-1}^{g}- \boldsymbol{w} \nabla_{\boldsymbol{x}_t}\left\|\Phi_\text{lp}\left(\boldsymbol{x}_{0}\right)-\Phi_\text{lp}\left(\hat{\boldsymbol{x}}^g_0\right)\right\|_2 +
\left\|\Phi_\text{hp}\left(\boldsymbol{x}_{0}\right) -\Phi_\text{hp}\left(\hat{\boldsymbol{x}}^g_0\right)\right\|_2. $}
\ENDFOR
 \end{algorithmic}
  \label{alg:supp_FDD_alg}
 \end{algorithm}
  
\begin{figure*} [htp!] \includegraphics[width=\textwidth]{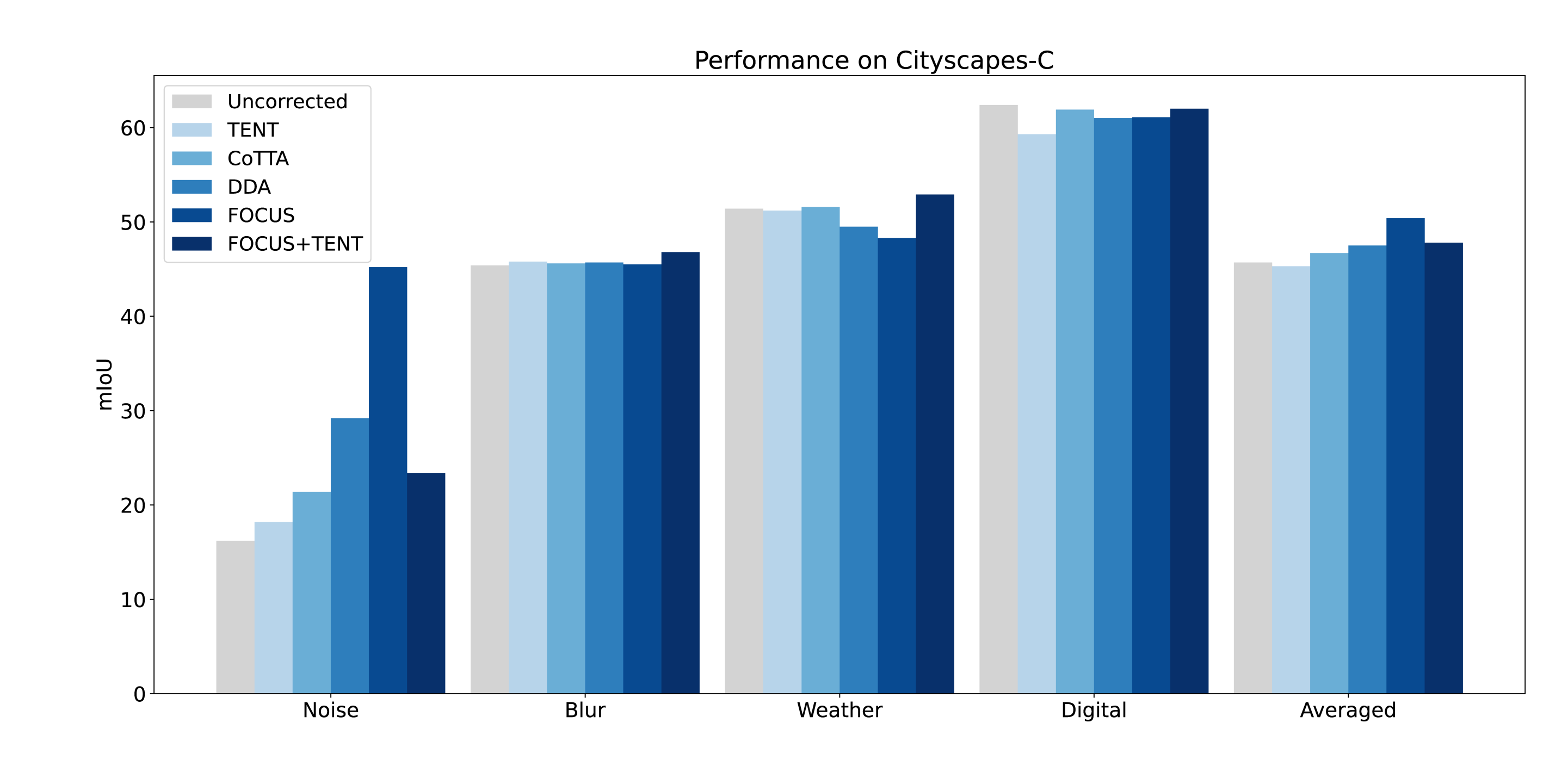}
\captionsetup{width=\textwidth}
\caption{\textbf{Quantitative results for Cityscapes-C}. FOCUS demonstrates the strongest performance (`Averaged') and demonstrates strong synergy (`FOCUS+TENT') for the corruption categories `Blur', `Weather' and `Digital'. 
Note that `FOCUS+TENT' demonstrates stronger averaged performance compared to 'TENT' across all 4 corruption categories.  Full per-corruption results in Table \ref{tab:TTA_Cityscapes_c_test}. }
\label{fig:Fig4_City-c_Quant}
\end{figure*}

\begin{figure*} [htp!] \includegraphics[width=\textwidth]{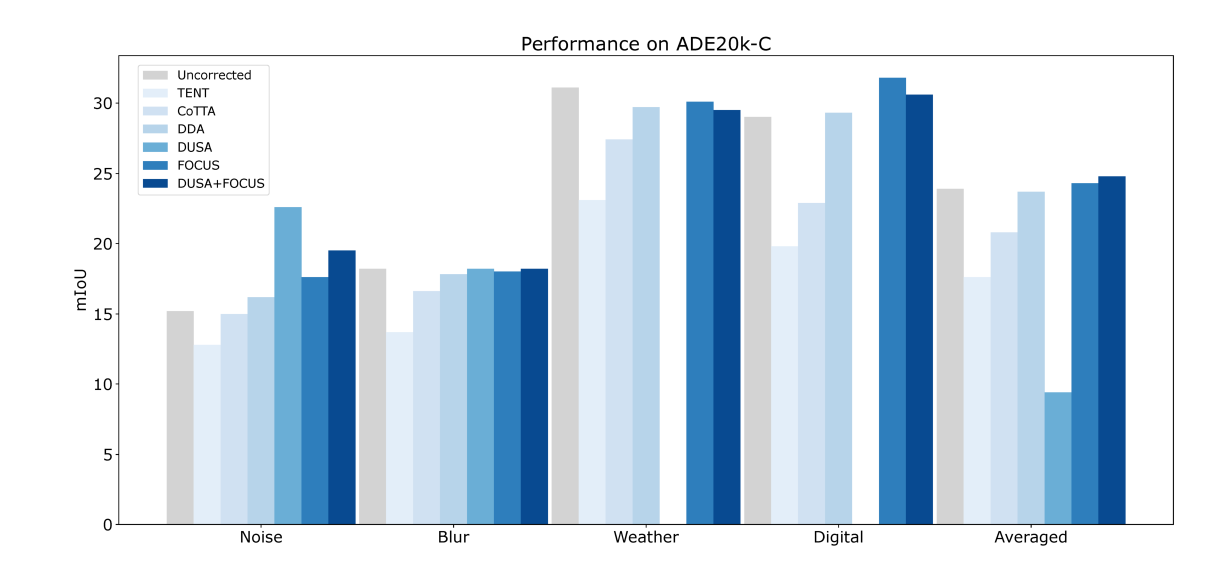}
\captionsetup{width=\textwidth}
\caption{\textbf{Quantitative results for ADE20k-C}. Performance for DUSA \cite{Li_DUSA_NeurIPS2024} starts strong initially, but degrades rapidly from catastrophic forgetting (`Weather', `Digital' ). FOCUS also demonstrates stronger performance compared to other model adaptation (\eg TENT, CoTTA) and input adaptation methods (DDA). Additionally, `FOCUS+DUSA' recovers the lost performance and even surpasses `FOCUS', suggesting the potential benefits of combining model adaptation methods with input adaptation methods.}
\label{fig:Fig5_ADE20k-c_Quant}
\end{figure*}

\subsection{Metrics}
We adopt the commonly used mean Intersection over Union(mIoU) and the Averaged Absolute Relative Error (Avg. Abs. Rel. Error) as the quantitative evaluation metrics for semantic segmentation and depth estimation respectively. 
A higher mIoU and a lower Avg. Abs. Rel. Error indicates stronger performance.

\subsection{Test-Time Adaptation Performance}
\textbf{Semantic Segmentation}
 Our method outperforms the baselines on averaged mIoU for Cityscapes-C (Figure \ref{fig:Fig4_City-c_Quant}) and ADE20k-C (Figure \ref{fig:Fig5_ADE20k-c_Quant}).
While DUSA \cite{Li_DUSA_NeurIPS2024} performs well initially, performance decreases rapidly after model fine tuning on a few corruptions.
We suggest that the reduced performance can be attributed to the following causes.
Firstly, the task model may overfit to test data, reducing the quality of the prediction logits when the test data distribution shifts significantly.
Since the diffusion model is conditioned on the prediction logits, the reduced quality of the prediction logits worsens the noise estimates, which sends erroneous supervisory signals to the task model and the diffusion model.
Additionally, since the diffusion model is also finetuned on the test data, some of the existing knowledge may be degraded as the diffusion model overfits to the test data.
These factors might account for the steep performance drop compared to CoTTA and TENT.

Our improved averaged performance relative to DDA, which uses predefined, image-level low frequency priors, indicates the importance of learned spatially adaptive priors.
Across Cityscapes and ADE20k, we observe stronger performance compared to DDA across nearly all corruption types except for `Weather' (Cityscapes-C). 

Generally, we observe reduced gains on `Blur' and `Weather' categories.
These corruptions cause signal loss via occlusions (`Frost',`Fog'), and degrade structural information via blurring.
For model adaptation methods (\eg TENT, CoTTA), performance is lower than the baseline `Uncorrected' for ADE20k-C (Figure \ref{fig:Fig5_ADE20k-c_Quant}) and comparable with the baselines `Uncorrected' for Cityscapes-C (Figure \ref{fig:Fig4_City-c_Quant}).
We suggest the reduced performance is caused by poorer pseudo label quality resulting from model overfitting to the earlier corruption category `Noise'.

Recent work \cite{luo2024controlling} introduce a noise-aware classifier to characterise the corruptions in the image, enabling the use of corruption-specific denoising techniques.
However, training such a classifier requires prior knowledge and is complex to implement.
Instead, we present a simple, straightforward solution to address these `antagonistic corruptions'.
The FOCUS denoised images provide an additional source of supervision for model adaptation methods \cite{WangSLOD21,Li_DUSA_NeurIPS2024} via the standard cross-entropy loss for semantic segmentation and consistency loss for depth estimation. 
We describe the additional loss functions in Section \ref{subsec:TTA_loss}.

Integrating input-adaptation methods such as FOCUS with model adaptation methods also yields synergistic effects.
We note that \textbf{FOCUS mitigates catastrophic forgetting for model adaptation based methods} such as TENT \cite{WangSLOD21} and DUSA \cite{Li_DUSA_NeurIPS2024}.
FOCUS improves adaptation performance for semantic segmentation (Fig. \ref{fig:Fig4_City-c_Quant},\ref{fig:Fig5_ADE20k-c_Quant}) and for depth estimation (Table \ref{tab:CTTA_NYU2k_c}), highlighting the synergistic effects of combining input and model adaptation. 

\begin{figure*}[ht]  \includegraphics[width=\textwidth]{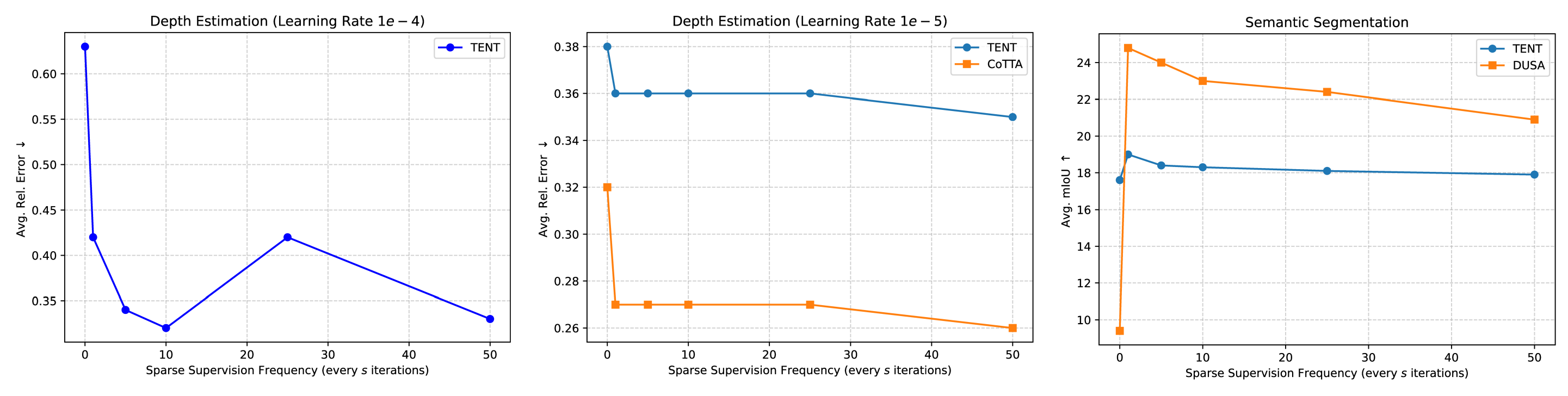}
\captionsetup{width=\textwidth}
\caption{Performance for sparse supervision using pseudo labels from FOCUS denoised images on  a) Depth Estimation (NYU2k-C, Learning Rate $1e-4$) b) Depth Estimation (NYU2k-C, Learning Rate $1e-5$) c) Semantic Segmentation (ADE20k-C). We provide additional supervision every $s$ iterations - /ie larger $s$ \textrightarrow more sparse supervision.}
\label{fig:sparse_sup}
\end{figure*}
TENT only updates the affine parameters instead of updating all parameters (CoTTA). 
This might make TENT less sensitive to changes in learning rate.
Nevertheless, FOCUS improves performance for both TENT and CoTTA (learning rate $10^{-4}$), highlighting the utility of combining input and model adaptation. 

\begin{figure*}  \includegraphics[width=\textwidth]{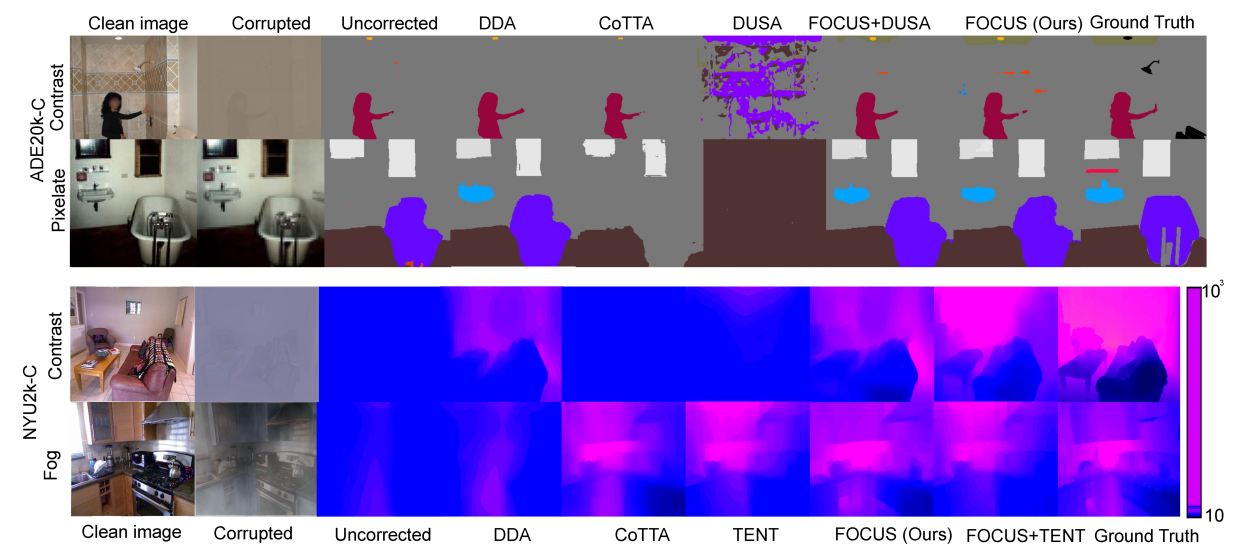}
\captionsetup{width=\textwidth}
\caption{Qualitative examples for semantic segmentation (ADE20k-C) and depth estimation (NYU2k-C). Full per-corruption results in Tables \ref{tab:TTA_NYU2k_c_test_1e4} and \ref{tab:TTA_NYU2k_c_test_1e5}. }
\label{fig:fig5_qual}
\end{figure*}

We also show some qualitative results in Fig. \ref{fig:fig5_qual}.
Importantly, model adaptation methods are heavily influenced by original task model performance.
For example, the source pretrained depth estimation model (`Uncorrected') performs poorly on the image degraded by `Contrast'. 
Consequently, both TENT and CoTTA perform poorly, underscoring the key requirement that strong out-of-distribution performance of the task model is essential for effective test time adaptation \cite{pmlr-v202-zhao23d},
In contrast, FOCUS is not as reliant on task model performance and provides additional improvements when combined with model adaptation methods (`FOCUS+TENT'). 

\subsection{Computational Costs}
\label{subsect:Compute_costs}
\begin{table}[h]
\centering
\caption{Computational cost comparison of adaptation methods for ADE20k-C. *: We add up the GPU memory costs from diffusion-driven denoising and fine tuning the task model.}
\label{tab:comp-cost}
\begin{tabularx}{\columnwidth}{lXXX}
\toprule
\textbf{Method} & \textbf{Trainable Params (M)} & \textbf{Time (s/frame)} & \textbf{Memory (GB)} \\
\midrule
Baseline     & 0     & 0.12  & $\sim$ 2  \\
TENT         & 0.4   & 0.18  & $\sim$ 6  \\
CoTTA        & 81.4  & 2.64  & $\sim$ 12 \\
DDA          & 0     & 15.58 & $\sim$ 24 \\
DUSA         & 1300  & 4.08  & $\sim$ 65 \\
FOCUS        & 27.5  & 17.72 & $\sim$ 24 \\
FOCUS+DUSA   & 109   & 19.21 & $\sim$ 44* \\
\bottomrule
\end{tabularx}
\end{table}
Training Y-FPN requires around 15GB GPU memory.
For FOCUS, we use the ImageNet pretrained diffusion model  without any additional finetuning.
Running the diffusion-driven diffusion denoising requires around 32GB of GPU memory (NVIDIA L40 46GB) for a batch size of 1 for an input image resolution $1024\times512$ pixels (Cityscapes-C), around 24GB for $512\times512$ pixels (ADE20k-C)
and about 28GB for $640\times512$ pixels (NYU2k-C).

Our work attempts to strike a balance between performance and minimizing computational costs.
The computational costs at test time in terms of GPU memory are higher than conventional test time adaptation methods (TENT, CoTTA), but less than those of DUSA (Table \ref{tab:comp-cost}).
Since we do not fine-tune the diffusion model for FOCUS+DUSA, we only require around 20GB GPU memory for model adaptation.
Combined with the GPU memory required during FOCUS diffusion-driven denoising, we use a total of approximately 44GB of GPU memory, while DUSA requires around 65GB of GPU memory at test time.

However, since inference speeds for diffusion-driven methods pose a bottleneck, we study the effect of intermittent supervision on adaptation performance in the following section (Section \ref{subsec:Focus_supervision_freq}).

\textbf{Depth Estimation} 
\begin{table}[ht]
\centering
\caption{Quantitative results (Averaged Absolute Relative error $\downarrow$) for depth estimation (NYU2k-C). }
\begin{tabularx}{\columnwidth}{lXX}
\toprule
\textbf{Method} & \textbf{LR = $10^{-4}$} & \textbf{LR = $10^{-5}$} \\
\midrule
Uncorrected & 0.710 & 0.710 \\
TENT & 0.625 & 0.378 \\
CoTTA & \textit{inf}\footnotemark & 0.321 \\
DDA & 0.693 & 0.693 \\
FOCUS  & 0.479 & 0.479 \\
FOCUS + TENT & \textbf{0.418} & 0.361 \\
FOCUS + CoTTA & \textit{inf}\footnotemark[\value{footnote}] & \textbf{0.269} \\
\bottomrule
\end{tabularx}
\label{tab:CTTA_NYU2k_c}
\end{table}
\footnotetext{Relative error diverged due to degenerate model output.}
FOCUS outperforms the benchmarks at learning rate $10^{-4}$ and demonstrates competitive performance at learning rate $10^{-5}$. in Table \ref{tab:CTTA_NYU2k_c}.
While TENT, CoTTA are sensitive to  learning rate changes, we show that FOCUS can stabilize performance for TENT (`FOCUS+TENT') across learning rates.
TENT only updates the affine parameters instead of updating all parameters (CoTTA). 
This might make TENT less sensitive to changes in learning rate.
Nevertheless, FOCUS improves performance for both TENT and CoTTA (learning rate $10^{-4}$), highlighting the utility of combining input and model adaptation. 
\subsection{Frequency of FOCUS-driven Supervision for Model Adaptation}
\label{subsec:Focus_supervision_freq}

To investigate the balance between adaptation effectiveness and inference efficiency, we studied the effect of intermittent supervision from the FOCUS denoised images on other model adaptation methods for depth estimation and semantic segmentation (Figure \ref{fig:sparse_sup}).
Specifically, we update the task model for the baselines (semantic segmentation on ADE20k-C: TENT \cite{WangSLOD21}, DUSA  \cite{Li_DUSA_NeurIPS2024} and depth estimation on NYU2k-C: TENT  \cite{WangSLOD21}, CoTTA \cite{Wang_cvpr2022_continual}) with additional supervision from the FOCUS denoised images every (1, 5, 10, 25, 50) iterations.
This design evaluates whether FOCUS can still regularize model adaptation despite limited supervision.

Our results on ADE20k-C reveal that while performance decreases as the frequency of supervision from FOCUS decreases, performance is still higher compared to model adaptation performance without supervision ($k$=0).
\textbf{We found that adaptation performance at lower supervision frequencies ($k=5,10$) is comparable with supervision at $k=1$ for both semantic segmentation and depth estimation tasks, suggesting the effectiveness of regularising model adaptation with pseudo labels from denoised images.}
Additionally, performance improved at $k$ = 50 (supervision from FOCUS denoised images is applied every 50 iterations) relative to $k$ = 0 , demonstrating the effectiveness of sparse FOCUS-driven supervision for model adaptation methods.

For depth estimation, we observe that additional supervision with FOCUS reduces the impact of increasing learning rate for model adaptation on depth estimation tasks.
Interestingly, we note for TENT and CoTTA, varying the frequency of FOCUS-driven supervision at learning rate $10^{-5}$ does not negatively impact performance.
Furthermore, at higher learning rates $10^{-4}$, we observe that reducing the frequency of FOCUS-driven supervision can even improve performance.
This suggests that applying denoised images, even intermittently and rarely, have a strong regularization effect for test-time model adaptation methods.

\subsection{Effect of the Frequency Loss Weight $\lambda_{2}$}
\label{subsec:freq_loss_weight}
\begin{table}[h]
    \centering
    \caption{Evaluation of frequency loss weights during Y-FPN training for semantic segmentation (Cityscapes-C). We use mIoU (\textuparrow) for the performance metrics. }
    \begin{tabularx}{\columnwidth}{lXXXX}
        \hline
        $\lambda_{2}$ & Gaussian Noise & Defocus Blur & Snow & Pixelate \\
        \hline
        0.01& 41.40 & \textbf{55.80} & 29.73& \textbf{74.49}\\
        0.1& 32.99& 54.24 & 23.02& 73.77 \\
        0.2 & \textbf{41.97} & 55.08	& \textbf{33.49}	&74.10 \\
           0.4 & 26.26 & 55.29 & 24.93 & 73.89\\
        \hline
    \end{tabularx}
    \label{tab:FrequencyLossweights_city_c}
    \end{table}
In this section, we evaluate the effect of $\lambda_{2}$ on adaptation performance (Table \ref{tab:FrequencyLossweights_city_c}).
We observe that $\lambda_{2}$ affects adaptation performance for `Noise' (`Gaussian Noise'), followed by `Weather' (`Snow') and `Blur'(`Defocus Blur') type corruptions. 
We found that optimal hyperparameter values lie within the range [0.1,0.2].

\subsection{Scale Weight \textit{w}}
\label{subsec:scale_weight}
\begin{table}[h]
\centering
\caption{Comparison of scale weights for semantic segmentation (Cityscapes-C). We use mIoU ($\uparrow$) for the performance metrics.}
\label{tab:scale_weights_city_c}
\begin{tabularx}{\columnwidth}{lXXXX}
\toprule
Scale weight & Gaussian Noise & Defocus Blur & Snow & Pixelate \\
\midrule
0.5 & 39.03 & 53.20 & 29.12 & 71.71 \\
1   & 40.70 & 54.56 & 31.26 & 73.19 \\
2   & 41.97 & \textbf{55.08} & \textbf{33.49} & \textbf{74.10} \\
4   & \textbf{42.33} & 54.76 & 30.03 & 73.80 \\
\bottomrule
\end{tabularx}
\end{table}

\begin{table}[h]
\centering
\caption{Comparison of scale weights for depth estimation (NYU2k-C). We use Averaged Absolute Relative Error ($\downarrow$) for the performance metrics.}
\label{tab:scale_weights_nyu2k_c}
\begin{tabularx}{\columnwidth}{lXXXX}
\toprule
Scale weight & Gaussian Noise & Defocus Blur & Snow & Pixelate \\
\midrule
1 & 0.34 & 1.40 & 0.75 & 0.40 \\
3 & \textbf{0.22} & 1.24 & \textbf{0.69} & 0.25 \\
6 & 0.22 & \textbf{1.07} & 0.71 & \textbf{0.16} \\
\bottomrule
\end{tabularx}
\end{table}
Scale weight is another important hyperparameter because it directly controls the conditioning strength of the frequency priors during diffusion-driven denoising.
\textbf{We found that increasing the scale weights generally improved performance for semantic segmentation (Table \ref{tab:scale_weights_city_c}) and depth estimation (Table \ref{tab:scale_weights_nyu2k_c}), which aligns with our expectation that preserving the semantic information during denoising }improved adaptation performance.

\subsection{Ablation Study}
We present ablation studies for ADE20k-C and NYU2k-C (Table \ref{tab:ablation_study_pair}).
\begin{table}[h]
\centering
\setlength{\tabcolsep}{5pt} % Adjust spacing for fit
\caption{Ablation study on ADE20k-C (semantic segmentation) and NYU2k-C (depth estimation). Metrics: mIoU (\textuparrow) and Abs. Rel. Error (\textdownarrow). Full details in Supplementary.}
\label{tab:booktabs_ablation}
\begin{tabularx}{\columnwidth}{lXX}
\toprule
Method & ADE20k-C (mIoU) $\uparrow$  & NYU2k-C (Abs. Rel. Err.) $\downarrow$ \\
\midrule
Baseline               & 23.9 & 0.710 \\
+ Y-FPN only           & 22.3 & 0.646 \\
\midrule
+ Unconditioned Diff.  & 22.2 & 0.669 \\
+ Low-Freq. Prior      & 23.8 & 0.599 \\
+ High-Freq. Prior     & 24.2 & 0.605 \\
\bottomrule
\end{tabularx}
\label{tab:ablation_study_pair}
\end{table}
We find that FOCUS achieves noticeable improvements compared to FOCUS (Y-FPN only), indicating that conditioned diffusion-driven denoising improves performance. 
Unconditioned diffusion-driven denoising, however, can degrade performance, further highlighting the importance of conditioning diffusion model during image denoising to preserve semantic information..
\noindent \textbf{Comparison of Frequency Priors} Conditioning with learned low frequency priors only improves performance for both semantic segmentation and depth estimation tasks.
Including high frequency priors further improves performance for semantic segmentation, but slightly reduces it for depth estimation.
\textbf{This suggests that conditioning with low frequency priors is generally sufficient for good performance, though we observed that for corruptions that degrade high frequency information (\eg `Contrast'), conditioning with high and low frequency priors improves performance.}
To further investigate the effect of frequency perturbations on model adaptability, we perform the comparison study in the following section (Section \ref{subsec:freq_perturb}).

\subsubsection{Effect of different Frequency Perturbations in FrequencyMix}
\label{subsec:freq_perturb}
\begin{table}[h]
\centering
\caption{Evaluation of frequency perturbations during Y-FPN training for FOCUS semantic segmentation (Cityscapes-C). We use mIoU ($\uparrow$) for the performance metrics.}
\label{tab:amp_perturb_city_c}
\begin{tabularx}{\columnwidth}{lXXXX}
\toprule
 & Gaussian Noise & Defocus Blur & Snow & Pixelate \\
\midrule
AugLow            & 28.8 & 54.4 & 25.0 & \textbf{75.2} \\
AugHigh           & 40.3 & \textbf{55.3} & 31.5 & 74.4 \\
AugRnd            & 31.4 & 54.7 & 22.0 & 72.2 \\
All perturbations & \textbf{42.0} & 55.1 & \textbf{33.5} & 74.1 \\
\bottomrule
\end{tabularx}
\end{table}
Since training data diversity is an important factor determining generalizability, we evaluate the effect of the 3 main frequency perturbations (Table \ref{tab:amp_perturb_city_c}).
We train 3 Y-FPN network variants by combining the base augmentations (uniform noise and uniform amplitude perturbations) with one of the 3 main frequency perturbations (AugLow, AugHigh and AugRnd) on Cityscapes.
We observe that AugHigh is more effective than AugLow or AugRnd in improving performance on `Gaussian Noise', `Defocus Blur' and `Snow'.
\textbf{While AugHigh has the strongest effect on performance compared to AugLow and AugRnd, performance is still lower than Y-FPN trained with all the frequency perturbations.}
In almost all corruptions, combining all frequency perturbations enhances performance beyond that of each of the frequency perturbations, suggesting the importance of including all the frequency perturbations during Y-FPN training.

\subsection{Number of timesteps  \textit{N}}
\label{subsec:timesteps}
\begin{table}[h]
    \centering
    \caption{Comparison of timesteps for semantic segmentation (Cityscapes-C). We use mIoU (\textuparrow) for the performance metrics.}
    \begin{tabularx}{\columnwidth}{lXXXX}
        \hline
        Timesteps & Gaussian Noise & Defocus Blur & Snow & Pixelate \\
        \hline
        1& 40.6& 56.1&32.3 &\textbf{74.6} \\
        5& 40.9& \textbf{56.1}& 33.2&74.3 \\
        15& \textbf{42.0} & 55.1&\textbf{33.5} &74.1 \\
        25& 40.9& 55.1&33.2 & 74.3\\
        50& 41.9& 55.2& 32.1& 73.9\\
        \hline
    \end{tabularx}
    \label{tab:cityc_Nsteps}
\end{table}
\begin{table}[h]
    \centering
    \caption{Comparison of timesteps for depth estimation (NYU2k-C). We use Averaged Absolute Relative Error (\textdownarrow) for the performance metrics.}
    \begin{tabularx}{\columnwidth}{lXXXX}
        \hline
        Timesteps & Gaussian Noise & Defocus Blur & Snow & Pixelate \\
        \hline
        1& 1.16& 1.26 & 0.82 & 0.15\\
        5 & 1.27 & 1.15 & 0.81 & 0.15 \\
        15 & 1.06 & 1.05 & 0.74& 0.15\\
        25 & 0.22 & 1.24 & 0.69 & 0.25 \\
        50 & 0.32 & 1.06 & 0.63 & 0.17 \\
        \hline
    \end{tabularx}
    \label{tab:NYUc_Nsteps}
\end{table}
For diffusion-driven denoising, timestep selection is important and also widely recognized as a critical hyperparameter for diffusion models.
We compare performance for semantic segmentation on Cityscapes-C (Table \ref{tab:cityc_Nsteps}) and depth estimation on NYU2k-C.
Our results for segmentation shows that increasing the number of reverse denoising timesteps reduces performance on `Blur' (\ie `Defocus Blur') but improves it for `Noise' (`Gaussian Noise').
The results show that the optimal number of time steps are within the range [15,25].
As seen from the variation across timesteps, the number of timesteps affects performance more noticeably for `Gaussian Noise',`Defocus Blur' and `Snow' compared to `Pixelate'.

\subsection{Clean Performance}
\begin{table}[h]
\centering
\setlength{\tabcolsep}{8pt} 
\caption{Performance on clean images for semantic segmentation (ADE20k-C) and depth estimation (NYU2k-C). Metrics: mIoU (\textuparrow) and Avg. Abs. Rel. Error (\textdownarrow).}
\label{tab:clean_performance}
\begin{tabularx}{\columnwidth}{lXX}
\toprule
\textbf{Method} & \textbf{ADE20k-C (mIoU)} $\uparrow$ & \textbf{NYU2k-C (Avg. Abs. Rel. Error)} $\downarrow$ \\
\midrule
Uncorrected & 49.1 & 0.162 \\
TENT        & 34.7 & 0.232 \\
CoTTA       & 46.7 & 0.174 \\
DDA         & 48.0 & 0.179 \\
DUSA        &  0.0 & -   \\
FOCUS       & 48.0 & 0.176 \\
\bottomrule
\end{tabularx}
\label{tab:clean_perf}
\end{table}
Since FOCUS is an image denoising approach, we wanted to evaluate performance on clean images (Table \ref{tab:clean_perf}).
For the model adaptation methods, we append an additional category `Clean' (comprising the original test images) to the existing 15 corruptions for ADE20k-C and NYU2k-C.
For semantic segmentation, FOCUS demonstrated stronger performance compared to `TENT', `CoTTA' and `DUSA' while showing similar clean performance with `DDA'.
For depth estimation, FOCUS demonstrated competitive performance with CoTTA.
\textbf{The results suggest that FOCUS preserves the much of the original semantic information for both semantic segmentation and depth estimation with minimal degradation, compared to model adaptation methods, which do not effectively adapt to the clean images and restore original performance.}
\subsection{Performance on Convolutional and Transformer based architectures }
\begin{figure}[h]
  \centering
\includegraphics[width=\columnwidth]{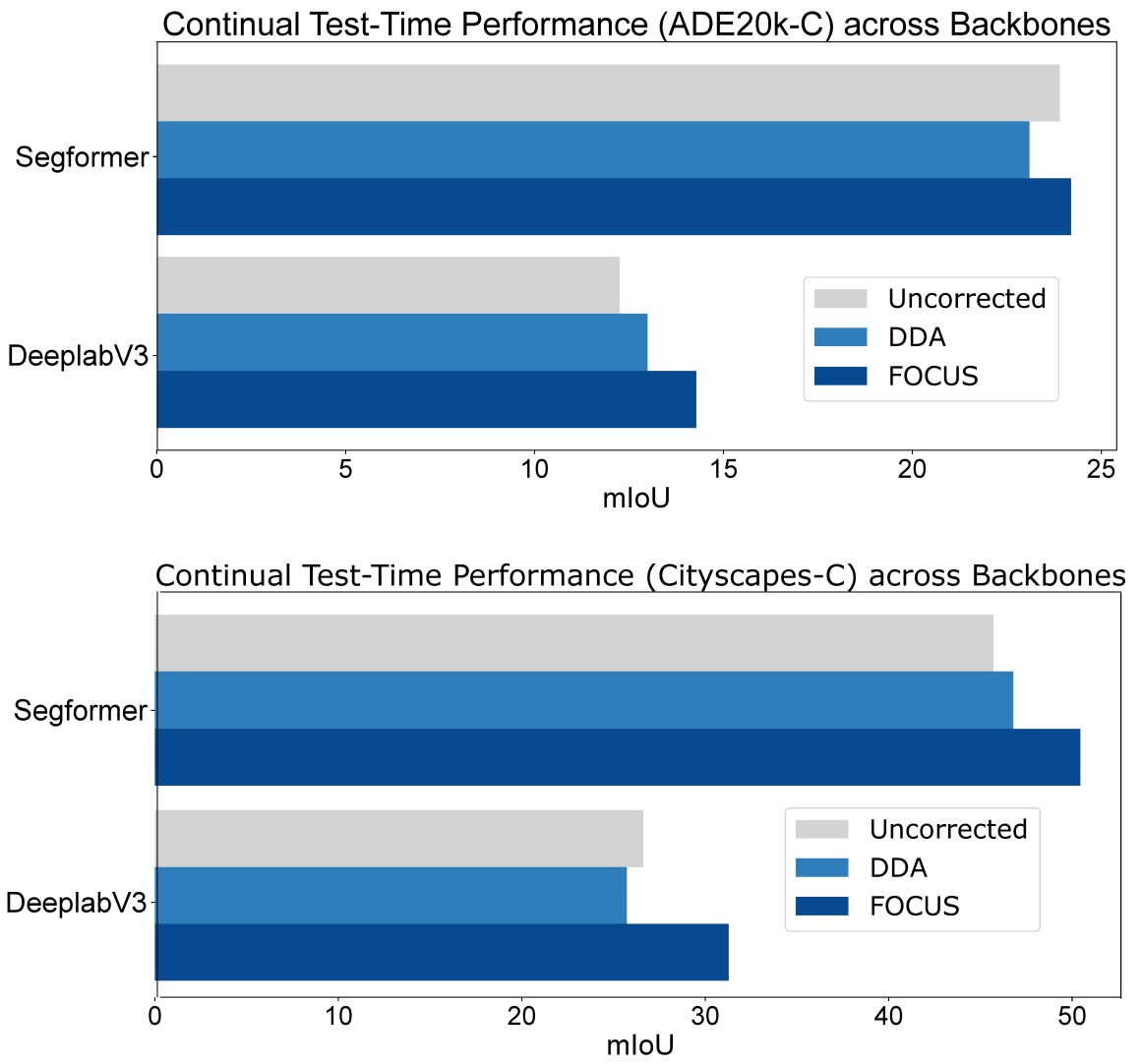}
 \captionsetup{width=\columnwidth}
\caption{Performance across transformer-based (Segformer) and convolutional-based (DeeplabV3-R101) architectures for ADE20k-C and Cityscapes-C.}
\label{fig:fig_modelag}
\end{figure}
FOCUS demonstrates consistent improvements over DDA and baseline uncorrected performance for both transformer and CNN-based architectures (Fig. \ref{fig:fig_modelag}), suggesting that our learned frequency priors are generalizable.

\subsection{Limitations}
While the Y-FPN layer itself is lightweight (with fast inference) and requires modest computational resources during training, FOCUS has longer inference times due to diffusion-driven denoising, which may limit its applicability for real-time test-time adaptation. To address this, we explored sparse, intermittent supervision using pseudo labels derived from FOCUS-denoised images (Section \ref{subsec:Focus_supervision_freq}). 

Our experiments demonstrate that intermittent regularization with denoised images improves adaptation performance, FOCUS synergizes effectively with model-adaptation methods. This suggests that combining both approaches maintains model plasticity by adapting to test data while preserving critical task-relevant information through fine-tuning with denoised inputs.

\section{Conclusion}
In this work, we introduce \textbf{F}requency \textbf{O}ptimized
\textbf{C}onditioning of Diff\textbf{US}ion (FOCUS),
a frequency-aware, diffusion-driven input adaptation framework for continual test-time adaptation in dense prediction. 
FOCUS conditions the denoising process on frequency priors extracted by a lightweight Y-shaped Frequency Prediction Network (Y-FPN) that is trained via a novel FrequencyMix augmentation strategy. 
Beyond strong standalone gains across diverse datasets and tasks, FOCUS enhances existing model adaptation methods by generating reliable pseudo-labels from the denoised inputs, which also mitigates catastrophic forgetting. 
We believe this work opens new directions for integrating generative models with frequency-aware conditioning to improve robustness and adaptability in dense prediction under real-world distribution shifts.

\clearpage
\bmhead{Supplementary information}
We include the full quantitative results for test-time adaptation performance for semantic segmentation (ADE20k-C and Cityscapes-C) and depth estimation (NYU2k-C).
\begin{sidewaystable*}
\centering
\caption{Test Time Adaptation Performance on Cityscapes-C (Semantic Segmentation). \textbf{Bold} and \underline{underline} denote best and runner-up results.  We also evaluate performance for  model adaptation with additional intermittent supervision (every $k$ iterations) with pseudo labels from the FOCUS denoised images.}
\resizebox{\textwidth}{!}{
\begin{tabular}{l *{16}{c}} 
\toprule
Time & \multicolumn{15}{c}{$t \xrightarrow{\hspace*{17cm}}$} \\
\toprule
\multirow{2}{*}{Method} & \multicolumn{3}{c}{Noise} & \multicolumn{4}{c}{Blur} & \multicolumn{4}{c}{Weather} & \multicolumn{4}{c}{Digital} & \multirow{2}{*}{Avg. mIoU ($\uparrow$)} \\
\cmidrule(lr){2-4} \cmidrule(lr){5-8} \cmidrule(lr){9-12} \cmidrule(lr){13-16}
& Gauss. & Shot & Impul. & Defoc. & Glass & Motion & Zoom & Snow & Frost & Fog & Brit. & Contr. & Elastic & Pixel & JPEG \\
\midrule
Segformer & 15.1 & 17.0 & 16.7 & \underline{56.7} & 44.3 & 56.0 & 24.6 & 25.7 & 28.5 & \textbf{72.7} & \underline{78.6} & \textbf{66.9} & \underline{77.8} & 68.0 & 37.1 & 45.7 \\
Tent (ICLR'21) & 17.5 & 19.7 & 17.5 & \textbf{56.8} & 45.5 & 56.0 & \textbf{25.0} & 27.0 & \textbf{30.5} & 70.2 & 77.3 & 64.8 & 76.6 & 65.4 & 30.7 & 45.3  \\
CoTTA (CVPR'22) & 19.2 & 23.7 & 21.4 & 56.5 & 44.9 & \underline{56.5} & 24.6 & 27.1 & 28.2 & \underline{72.2} & \textbf{79.0} & \underline{65.9} & \textbf{78.2} & 67.3 & 36.2 & 46.7 \\
 DDA & 16.5 & 16.8 & 16.3 & 22.7 & 18.4 & 20.5 & 9.1 & 19.6 & 18.6 & 32.7 & 39.4 & 19.8 & 24.4 & 36.1 & 35.7 & 23.1 \\
\rowcolor{lightpurple} FOCUS (Ours) &
\textbf{42.0} & \textbf{51.8} & \textbf{42.8} & 55.1 & \textbf{50.2} & 55.2 & 21.5 & \textbf{33.5} & 17.7 & 67.1 & 75.1 & 53.2 & 76.4 & \textbf{74.1} & \underline{40.9} & \textbf{50.4}\\
\midrule
\rowcolor{lightpurple} FOCUS (Ours) + TENT ($k$=1) &  19.3 & 25.4 & 25.6 & 56.2 & \underline{49.7} & \textbf{56.8} & \underline{24.7} & \underline{32.6} & \underline{29.5} & 71.0 & 78.4 & 57.8 & 77.7 & 69.9 & \textbf{42.7} & 47.8\\
\bottomrule
\end{tabular}
}
\label{tab:TTA_Cityscapes_c_test}
\end{sidewaystable*}

\begin{sidewaystable*}
\centering
\caption{Test Time Adaptation Performance on ADE20k-C (Semantic Segmentation). \textbf{Bold} and \underline{underline} denote best and runner-up results.  We also evaluate performance for  model adaptation with additional intermittent supervision (every $k$ iterations) with pseudo labels from the FOCUS denoised images.}
\resizebox{\textwidth}{!}{
\begin{tabular}{l *{16}{c}} 
\toprule
Time & \multicolumn{15}{c}{$t \xrightarrow{\hspace*{17cm}}$} \\
\toprule
\multirow{2}{*}{Method} & \multicolumn{3}{c}{Noise} & \multicolumn{4}{c}{Blur} & \multicolumn{4}{c}{Weather} & \multicolumn{4}{c}{Digital} & \multirow{2}{*}{Avg. mIoU ($\uparrow$)} \\
\cmidrule(lr){2-4} \cmidrule(lr){5-8} \cmidrule(lr){9-12} \cmidrule(lr){13-16}
& Gauss. & Shot & Impul. & Defoc. & Glass & Motion & Zoom & Snow & Frost & Fog & Brit. & Contr. & Elastic & Pixel & JPEG \\
\midrule
Segformer & 14.2 & 15.8 & 15.6 & \textbf{23.1} & 16.8 & \textbf{22.5} & \textbf{10.3} & \textbf{22.3} & \textbf{21.5} & \textbf{38.6} & \textbf{42.0} & \underline{23.1} & \textbf{24.5} & 33.1 & \underline{35.3} & \underline{23.9} \\
Tent (ICLR’21) & 11.2 & 13.6 & 13.6 & 16.8 & 14.0 & 17.0 & 6.8 & 17.5 & 17.0 & 27.5 & 30.5 & 15.7 & 18.2 & 22.4 & 22.9 & 17.6  \\
CoTTA (CVPR’22) & 14.6 & 15.7 & 14.7 & 21.9 & 15.2 & 20.6 & 8.5 & 18.7 & 17.3 & 34.8 & 38.9 & 15.0 & 20.5 & 27.8 & 28.4 & 20.8 \\
 DDA (CVPR’23) & 16.5 & 16.8 & 16.3 & \underline{22.7} & \underline{18.4} & 20.5 & 9.1 & 19.6 & 18.6 & 32.7 & 39.4 & 19.8 & 24.4 & \textbf{36.1} &  35.7 & 23.1 \\
DUSA (NeurIPS'24) & \textbf{21.2} & \textbf{24.0} & \textbf{22.7} & 21.7 & \textbf{24.0} & 20.2 & 7.1 & 0.0 & 0.0 & 0.0 & 0.0 & 0.0 & 0.0 & 0.0 & 0.0 & 9.4  \\
\rowcolor{lightpurple} FOCUS (Ours) & 17.1 &18.0 &17.7 & 22.5 & 17.3 & \underline{21.9} & \underline{10.1} & \underline{20.1} & \underline{20.3} & \underline{38.3} & 41.6 & 24.6 & \underline{24.3} & \underline{35.0} & 35.7 & 24.3 \\
\rowcolor{lightpurple} FOCUS (Ours) + DUSA ($k$=1)&  18.5 & \underline{20.7} & \underline{19.4} & 22.2 & \underline{18.4} & \underline{21.9} & \underline{10.1} & 19.4 & 19.3 & 37.5 & \underline{41.8} & \textbf{25.9} & 24.1 & \textbf{36.1} & \textbf{36.4} & \textbf{24.8}  \\
\rowcolor{lightpurple} FOCUS (Ours) + DUSA ($k$=5)   & \underline{19.2} & 19.9 & 18.2 & 21.0 & 17.4 & 21.1 & 9.6 & 19.7 & 18.7 & 37.0 & 41.0 & 25.6 & 22.3 & 34.6 & 35.3 & 24.0 \\
\bottomrule
\end{tabular}
}
\label{tab:TTA_ADE20k_c_test}
\end{sidewaystable*}

\begin{sidewaystable*}
\centering
\caption{Test Time Adaptation Performance on NYU2k-C (Depth Estimation, Learning Rate $10^{-4}$). \textbf{Bold} and \underline{underline} denote best and runner-up results.  We also evaluate performance for  model adaptation with additional intermittent supervision (every $k$ iterations) with pseudo labels from the FOCUS denoised images. CoTTA diverged to degenerate solutions.}
\resizebox{\textwidth}{!}{
\begin{tabular}{l *{16}{c}} 
\toprule
Time & \multicolumn{15}{c}{$t \xrightarrow{\hspace*{17cm}}$} \\
\toprule
\multirow{2}{*}{Method} & \multicolumn{3}{c}{Noise} & \multicolumn{4}{c}{Blur} & \multicolumn{4}{c}{Weather} & \multicolumn{4}{c}{Digital} & \multirow{2}{*}{Avg. Rel. Error \textdownarrow} \\
\cmidrule(lr){2-4} \cmidrule(lr){5-8} \cmidrule(lr){9-12} \cmidrule(lr){13-16}
& Gauss. & Shot & Impul. & Defoc. & Glass & Motion & Zoom & Snow & Frost & Fog & Brit. & Contr. & Elastic & Pixel & JPEG \\
\midrule
DenseDepth & 1.27 & 1.20 & 1.24 & 1.04 & 0.63 & \underline{0.65} & \underline{0.56} & 0.82 & 0.91 & \textbf{0.36} & \textbf{0.21} & 1.12 & \textbf{0.28} & \textbf{0.15} & 0.21 & 0.71 \\
Tent (ICLR'21) & 0.53 & 0.47 & \underline{0.42} & \underline{0.64} & \underline{0.59} & 0.75 & 0.93 & \underline{0.51} & 0.61 & 0.57 & 0.54 & 0.93 & 0.54 & 0.60 & 0.75 & 0.63  \\
DDA (CVPR'23) & 1.30 & 1.21 & 1.29 & 1.08 & 0.78 & 0.81 & 0.88 & 0.82 & 0.78 & 0.77 &  \underline{0.23} & 1.40 & 0.27 & 0.15 & \underline{0.19} & 0.80 \\
 \rowcolor{lightpurple} FOCUS (Ours) & \textbf{0.22} & \textbf{0.24} & \textbf{0.19} & 1.07 & 0.70 & 0.71 & 0.70 & 0.71 & \textbf{0.38} & 0.55 & \underline{0.23} & \underline{0.88} & 0.29 & \underline{0.16} & \textbf{0.17} & \underline{0.48}\\
  \rowcolor{lightpurple} FOCUS (Ours) + TENT($k$=1) & \underline{0.43} & \underline{0.38} & 0.44 & \textbf{0.53} & \textbf{0.50} & \textbf{0.47} & \textbf{0.40} & \textbf{0.40} & \underline{0.43} & \underline{0.38} & 0.31 & \textbf{0.64} &\underline{0.32} & 0.29 & 0.34 & \textbf{0.42} \\
\bottomrule
\end{tabular}
}
\label{tab:TTA_NYU2k_c_test_1e4}
\end{sidewaystable*}

 \begin{sidewaystable*}
\centering
\caption{Test Time Adaptation Performance on NYU2k-C (Depth Estimation, Learning Rate $10^{-5}$). \textbf{Bold} and \underline{underline} denote best and runner-up results.  We also evaluate performance for  model adaptation with additional intermittent supervision (every $k$ iterations) with pseudo labels from the FOCUS denoised images.}
\resizebox{\textwidth}{!}{
\begin{tabular}{l *{16}{c}} 
\toprule
Time & \multicolumn{15}{c}{$t \xrightarrow{\hspace*{17cm}}$} \\
\toprule
\multirow{2}{*}{Method} & \multicolumn{3}{c}{Noise} & \multicolumn{4}{c}{Blur} & \multicolumn{4}{c}{Weather} & \multicolumn{4}{c}{Digital} & \multirow{2}{*}{Avg. Rel. Error \textdownarrow} \\
\cmidrule(lr){2-4} \cmidrule(lr){5-8} \cmidrule(lr){9-12} \cmidrule(lr){13-16}
& Gauss. & Shot & Impul. & Defoc. & Glass & Motion & Zoom & Snow & Frost & Fog & Brit. & Contr. & Elastic & Pixel & JPEG \\
\midrule
DenseDepth & 1.27 & 1.20 & 1.24 & 1.04 & 0.63 & 0.65 & 0.56 & 0.82 & 0.91 & 0.36 & \underline{0.21} & 1.12 & 0.28 & \textbf{0.15} & 0.21 & 0.71 \\
Tent (ICLR'21) &  0.50 & 0.32 & 0.35 & 0.50 & 0.37 & 0.43 & 0.40 & 0.39 & 0.39 & 0.33 & 0.25 & 0.68 & 0.25 & 0.24 & 0.27 & 0.38 \\
CoTTA (CVPR'22)& 0.41 & 0.28 & 0.36 & 0.51 & \underline{0.29} & \underline{0.33} & \underline{0.30} & \underline{0.37} & 0.39 & \underline{0.22} & \textbf{0.17} & \underline{0.66} & \textbf{0.16} & 0.18 & 0.20 & \underline{0.32} \\
DDA (CVPR'23) & 1.30 & 1.21 & 1.29 & 1.08 & 0.78 & 0.81 & 0.88 & 0.82 & 0.78 & 0.77 & 0.23 & 1.40 & 0.27 & 0.15 & \underline{0.19} & 0.80 \\
 \rowcolor{lightpurple} FOCUS (Ours) & \textbf{0.22} & \underline{0.24} & \textbf{0.19} & 1.07 & 0.70 & 0.71 & 0.70 & 0.71 & \textbf{0.38} & 0.55 & 0.23 & 0.88 & 0.29 & \underline{0.16} & \textbf{0.17} & \underline{0.48}\\
\rowcolor{lightpurple} FOCUS (Ours) + TENT($k$=1) &0.41 & 0.32 & 0.35 & \underline{0.44} & 0.38 & 0.42 & 0.37 & \underline{0.37} & 0.39 & 0.33 & 0.25 & \textbf{0.63} & 0.25 & 0.24 & 0.28 & 0.36\\
  \rowcolor{lightpurple} FOCUS (Ours) + CoTTA ($k$=1) & \underline{0.28} & \textbf{0.23} & \underline{0.24} & \textbf{0.31} & \textbf{0.24} & \textbf{0.31} & \textbf{0.27} & \textbf{0.29} & \underline{0.29} & \textbf{0.20} & \textbf{0.17} & 0.71 & \underline{0.17} & \underline{0.16} & 0.20 & \textbf{0.27} \\
 \bottomrule
\end{tabular}
}
\label{tab:TTA_NYU2k_c_test_1e5}
\end{sidewaystable*}

\bmhead{Acknowledgements}
This research is supported by the National Research Foundation, Singapore under its AI Singapore Programme (AISG Award No: AISG4-GC-2023-008-1B).

\section*{Declarations}

\clearpage

\bibliography{main.bib}
\end{document}